\documentclass[mnsc,nonblindrev]{informs3_hide}
\OneAndAHalfSpacedXI 

\usepackage[english]{babel}
\usepackage[autostyle, english = american]{csquotes}
\MakeOuterQuote{"}
\usepackage[colorlinks,linkcolor=blue, citecolor=blue]{hyperref}
\usepackage[normalem]{ulem}

\usepackage{cleveref}
\crefname{subsection}{section}{subsections}
\usepackage{eqnarray}

\RequirePackage{algorithm}
\RequirePackage{algorithmic}
\usepackage{amsmath, bbm, enumitem, xparse, bm, graphicx, lipsum}
\usepackage{amssymb}
\usepackage{subcaption}

\newcommand{\eps}{\varepsilon}
\newcommand{\bI}{\mathbbm{1}}

\newcommand{\ALG}{\mathsf{ALG}}
\newcommand{\ALGD}{\mathsf{ALG}_{\mathsf{Dual}}}
\newcommand{\OPT}{\mathsf{OPT}}
\newcommand{\hOPT}{\hat{\mathsf{OPT}}}
\newcommand{\OPTInner}{\mathsf{OPT}}
\newcommand{\OPTOuter}{\mathsf{OPT}^{\mathsf{Covering}}}
\newcommand{\LInner}{\mathsf{L}}
\newcommand{\LOuter}{\mathsf{L}^{\mathsf{Covering}}}
\newcommand{\bLInner}{\bar{L}}
\newcommand{\bLOuter}{\bar{L}^{\mathsf{Covering}}}
\newcommand{\hLInner}{\hat{L}}

\newcommand{\Reg}{\mathsf{Regret}}

\newcommand{\bc}{\bm{c}}
\newcommand{\hbc}{\hat{\bm{c}}}

\newcommand{\tbc}{\tilde{\bm{c}}}

\newcommand{\bx}{\bm{x}}
\newcommand{\tbx}{\tilde{\bm{x}}}
\newcommand{\hbx}{\hat{\bm{x}}}

\usepackage[dvipsnames]{xcolor}

\usepackage{xparse}

\NewDocumentEnvironment{myproof}{o}
{\IfNoValueTF{#1}{\paragraph{{Proof.} }} {\paragraph{{#1.} }} }
{\hfill$\Halmos$}

\usepackage{natbib,bbm,multirow,multicol}
 \bibpunct[, ]{(}{)}{,}{a}{}{,}%
 \def\bibfont{\small}%
\TheoremsNumberedThrough     
\ECRepeatTheorems

\EquationsNumberedThrough    

\begin{document}


\RUNAUTHOR{Jiang}

\RUNTITLE{Constrained Online Two-stage Stochastic Optimization}

\TITLE{
\Large Constrained Online Two-stage Stochastic Optimization:\\ Near Optimal Algorithms via Adversarial Learning
}

\ARTICLEAUTHORS{%
\AUTHOR{$\text{Jiashuo Jiang}^{\dagger}$}

\AFF{\  \\
$\dagger~$Department of Industrial Engineering \& Decision Analytics, Hong Kong University of Science and Technology
}
}

\ABSTRACT{
We consider an online two-stage stochastic optimization with long-term constraints over a finite horizon of $T$ periods. At each period, we take the first-stage action, observe a model parameter realization and then take the second-stage action from a feasible set that depends both on the first-stage decision and the model parameter. We aim to minimize the cumulative objective value while guaranteeing that the long-term average second-stage decision belongs to a set. We develop online algorithms for the online two-stage problem from adversarial learning algorithms. Also, the regret bound of our algorithm cam be reduced to the regret bound of embedded adversarial learning algorithms. Based on our framework, we obtain new results under various settings. When the model parameter at each period is drawn from identical distributions, we derive \textit{state-of-art} $O(\sqrt{T})$ regret that improves previous bounds under special cases. Our algorithm is also robust to adversarial corruptions of model parameter realizations. When the model parameters are drawn from unknown non-stationary distributions and we are given machine-learned predictions of the distributions, we develop a new algorithm from our framework with a regret $O(W_T+\sqrt{T})$, where $W_T$ measures the total inaccuracy of the machine-learned predictions.
}



\maketitle

\section{Introduction}\label{sec:intro}
Stochastic optimization is widely used to model the decision making problem with uncertain model parameters. In general, stochastic optimization aims to solve the problem with formulation $\min_{\bc\in\mathcal{C}} \mathbb{E}_{\theta}[F_{\theta}(\bc)]$, where $\theta$ models parameter uncertainty and we optimize the objective on average. An important class of stochastic optimization models is the \textit{two-stage model}, where the problem is further divided into two stages. In particular, at the first stage, we decide the first-stage decision $\bc$ without knowing the exact value of $\theta$. At the second stage, after a realization of the uncertain data becomes known, an optimal second stage decision $\bx$ is made by solving an optimization problem parameterized by both $\bc$ and $\theta$, where one of the constraint can be formulated as $x\in\mathcal{B}(c, \theta)$. Here, $F_{\theta}(\bc)$ denotes the optimal objective value of the second-stage optimization problem. Two-stage stochastic optimization has numerous applications, including  transportation, logistics, financial instruments, and supply chain, among others \citep{birge2011introduction}.

In this paper, we focus on an ``online'' extension of the classical two-stage stochastic optimization over a finite horizon of $T$ periods. Subsequently, at each period $t$, we first decide the first-stage decision $\bc_t\in\mathcal{C}$, then observe the value of model parameter $\theta_t$, which is assumed to be drawn from an \textit{unknown} distribution $P_t$, and finally decide the second-stage decision $\bx_t$. In addition to requiring $\bx_t$ belonging to a constraint set parameterized by $\bc_t$ and $\theta_t$, we also need to satisfy a long-term global constraint of the following form: $\frac{1}{T}\cdot\sum_{t=1}^{T}\bx_t\in\mathcal{B}(\bm{C},\bm{\theta})$, where $\mathcal{B}(\bm{C}, \bm{\theta})$ is a set that depends on the entire sequence $\bm{\theta}=(\theta_t)_{t=1}^T$ and $\bm{C}=(\bc_t)_{t=1}^T$. We aim to optimize the total objective value over the entire horizon, and we measure the performance of our online policy by ``regret'', the additive loss of the online policy compared to the \textit{optimal dynamic policy} which is aware of $P_t$ for each period $t$.

Indeed, one motivating example for our model is the resource allocation problems that arises from many applications. Usually, the online decision making in the resource allocation problem is composed of two layers. The first one is the \textit{budget} layer, and the second one is the \textit{allocation} layer. At each period, the decision maker needs to decide a budget to be consumed within this period, which is the first-stage deicion $\bc$, and then, after the budget has been decided, the decision maker needs to decide how much resource, within the budget, to be allocated to fulfill the demand, which is the second-stage decision $\bx$. It is common in practice that over the entire horizon, the total allocated resource cannot surpass the initial capacities of the resources, or during the entire horizon, the cumulated satisfied demand cannot be smaller than certain thresholds, which represents the service-level constraints (e.g. \cite{hou2009theory}) or the fairness constraints (e.g. \cite{kearns2018preventing}). These operational requirements induce long-term constraints into our model.


Our online model is a natural synthesis of several widely studied models in the existing literature. Roughly speaking, we classify previous models on online learning/optimization into the following two categories: the \textit{bandits-based} model and the \textit{type-based} model. For the bandits-based model, we make the decision and then observe the (possibly stochastic) outcome, which can be adversarially chosen. The representative problems include multi-arm-bandits (MAB) problem, bandits-with-knapsacks (BwK) problem, and the more general online convex optimization (OCO) problem. For the type-based model, at each period, we first observe the type of the arrival, and we are clear of the possible outcome for each action (which can be type-dependent), and then we decide the action without knowing the type of future periods. Note that in the type-based model, we usually have a global constraint such that the cumulative decision over the entire horizon belongs to a set (otherwise the problem becomes trivial, just select the myopic optimal action at each period), which corresponds to the long-term constraint in our model. The representative problems for the type-based model include online allocation problem and a special case online packing problem where the objective function is linear and $\mathcal{B}(\bm{C}, \bm{\theta})$ is a polyhedron. We review the literature on these problems in \Cref{sec:RelatedWork}. For our model, if we assume the second-stage decision step is de-activated, i.e., at each period $t$, the second-stage decision $\bx_t$ is fixed as long as the first-stage decision $\bc_t$ is determined and the model uncertainty $\theta_t$ is realized, then, our model reduces to the bandits-based model described above. On the other hand, if the first-stage decision step is de-activated, i.e., $\mathcal{C}$ is a set of a fixed point, then, our model reduces to the type-based model described above.

From a different perspective, our online problem is also motivated from the computational challenge faced by existing approaches for the classical two-stage stochastic optimization problem. By examining the literature, most approaches for two-stage stochastic optimization, i.e., sample average approximation method \citep{shapiro2000rate} and stochastic approximation method \citep{nemirovski2009robust}, would require to solve the second-stage optimization problem for one or multiple times at every iteration, where the constraint $\bx_t\in\mathcal{B}(\bc_t, \theta_t)$ is involved. However, when the set $\mathcal{B}(\bc, \theta)$ possesses complex structures, the second-stage optimization step can be computationally burdensome. Therefore, alternatively, instead of requiring $\bx_t\in\mathcal{B}(\bc_t, \theta_t)$ for each $t\in[T]$ in our online problem, we require only the long-run average constraint $\frac{1}{T}\cdot\sum_{t=1}^{T}\bx_t\in\mathcal{B}(\bm{C}, \bm{\theta})$ is satisfied.  In other words, we allow the decision maker to violate the constraint set $\mathcal{B}(\bm{c}_t, \theta_t)$ for some rounds, but the entire sequence of decisions must fit the constraints at the very end. In this way, we are trading the regret for the online problem for computational efficiency of the classical two-stage problem, which is analogous to trading regret for efficiency for OCO problem \citep{mahdavi2012trading}.

\subsection{Our Approach and Results}\label{sec:OurResults}

Our main approach relies on regarding one long-term constraint as an \textit{expert}, and we reduce the problem of satisfying the long-term constraints as a procedure for finding the ``best'' expert. To be specific, at each period $t$, we assign a probability $\alpha_{i,t}$ to each long-term constraint $i$ that describes the set $\mathcal{B}(\bm{C}, \bm{\theta})$, where the summation of $\alpha_{i,t}$ over $i$ equals $1$ for each $t$. We show that if we are applying an adversarial learning algorithm to tackle the expert problem induced by satisfying the long-term constraints, which implies the rule to update $\alpha_{i,t}$, then the gap of constraints violation can be bounded by the regret bound of the adversarial learning algorithm. On the other hand, if we let $\lambda_{i,t}=\mu\cdot\alpha_{i,t}$ where $\mu$ is some scaling factor, then we can regard $\lambda_{i,t}$ as the \textit{Lagrangian dual variable} for the long-term constraint $i$ and we can apply another adversarial learning algorithm to minimize the accumulated Lagrangian value over the entire horizon, where the decision variable now is the first-stage decision. After both the first-stage decision and the Lagrangian dual variable are determined for the current period, we show that it is sufficient to decide the second-stage decision by solving a simple optimization problem. Meanwhile, the solved second-stage decision also serves as \textit{stochastic outcome} as input to the two adversarial learning algorithms, which help us update $\alpha_{i,t}$ and the first-stage decision for the next period. Such a primal-dual framework has also been developed in \cite{roth2016watch, rivera2018online, immorlica2019adversarial, castiglioni2022online, castiglioni2022unifying} for bandit optimization where the second-stage decision is de-activated and in \cite{gupta2014experts} for packing problem where the first-stage actions is de-activated. However, this the first time the primal-dual approach is developed for the two-stage problem. We now elaborate more on our results and compare with the previous primal-dual approach developed in the literature.

We consider the stationary setting as a starter, where the distribution $P_t$ is identical for each $t$. By applying the above framework, we derive an online algorithm, which we name \textit{Doubly Adversarial Learning} (DAL) algorithm, that satisfies the long-term constraints and achieve a regret bound of $O((G+F)\cdot\sqrt{T})+O(\sqrt{T\cdot\log m})$, where $G$ denotes the diameter of the feasible set for the first-stage actions, $F$ is a constant depending on the gradients of the objective function and the constraint functions that define the set $\mathcal{B}$, and $m$ denotes the number of constraints that characterize $\mathcal{B}$. This the first sublinear regret bound for the constrained online two-stage stochastic optimization. Compared to the previous primal-dual algorithms in the literature, our algorithm is different. For example, for bandit optimization problem where the second-stage decision is de-activated, as in \cite{castiglioni2022online} and \cite{castiglioni2022unifying}, the dual is restricted to a bounded set that is parameterized by the problem setting, whereas in our algorithm, the $L_1$ norm of the dual is naturally bounded by 1 since the dual specifies a distribution over the long-term constraints (regarded as experts in our approach). Moreover, the feedback for each expert in our algorithm is derived from a modified Lagrangian value (with formulation given in \eqref{eqn:defOutISP}), whereas the feedback for the dual variable in \cite{castiglioni2022online, castiglioni2022unifying} are defined as the Lagrangian value itself. The existence of the second-stage decision would make our update more complicated as the second-stage decision needs to be carefully selected to serve as unbiased estimator of the Lagrangian value. Finally, the previous primal-dual approaches for bandit optimization \citep{immorlica2019adversarial, castiglioni2022online, castiglioni2022unifying} are all for packing style long-term constraints. We show how our approach can be extended to deal with covering style long-term constraints, with details specified in \Cref{sec:Covering}.


We show that our DAL algorithm is robust to adversarial corruptions of the model parameters, i.e., the model parameter $\theta_t$ is corrupted to $\theta_t^c$ by an adversary, at each period. The adversarial corruption to a stochastic model can arise from the non-stationarity of the underlying distributions $\bm{P}$ \citep{jiang2020online}, or malicious attack and false information input to the system \citep{lykouris2018stochastic}. And the existence of the adversarial corruptions would make our problem deviate from the stationary setting and become a non-stationary setting where $P_t$ can be non-homogeneous for each $t$. We show that our online algorithm can tolerate a certain amount of corruptions and still achieve sublinear regret bound. We first show that if the total number of corruptions is $W$, then no online algorithm can achieve a regret bound better than $\Omega(W)$. We then show that our DAL algorithm achieves the regret bound $O(W+\sqrt{T})$ in the presence of adversarial corruptions, which matches the lower bound $\Omega(W)$. The linear dependency on $W$ of our DAL algorithm corresponds to the linear dependency established in \cite{gupta2019better} for the MAB problem.

Finally, we explore whether we can improve the regret bound $O(W+\sqrt{T})$ when $P_t$ is non-homogeneous for $t$, with the help of possible additional information. In practice, historical data are usually available for the distribution $P_t$. Therefore, we consider a prediction setting where we have a machine-learned prediction $\hat{P}_t$ for $P_t$ at each time period. In this case, we show that $W$ in the lower bound $\Omega(W)$ can now be replaced by the total inaccuracy of the predictions, i.e., the accumulated gap between $\hat{P}_t$ and $P_t$ for all $t$. In terms of the online algorithm, note that since we have prior predictions, there is no need to apply adversarial learning algorithm to decide the first-stage action. Instead, we can simply solve a two-stage stochastic optimization problem at each period to decide both the first-stage decision and the second-stage decision. The two-stage problem is formulated differently for different periods to handle the non-stationary of the underlying distributions. The formulation relies on the predictions, and as a result, our algorithm, which we name \textit{Informative Adversarial Learning} (IAL) algorithm, naturally combines the information provided by the predictions into the adversarial learning algorithm of the dual variables for the long-term constraints. Our IAL algorithm achieves a regret bound $O(W+\sqrt{T})$, which matches the lower bound $\Omega(W)$.

\subsection{Other Related Literature}\label{sec:RelatedWork}
Our model synthesizes the bandits-based model, where the representative problems include MAB problem, BwK problem and the more general OCO problem, and the feature-based model, where the representative problems include online allocation problem and a special case online packing problem. We now review these literature.

One representative problem for the bandits-based model is the BwK problem which reduces to MAB problem if no long-term constraints. The previous BwK results have focused on a stochastic setting \citep{badanidiyuru2013bandits, agrawal2014bandits}, where a $O(\sqrt{T})$ regret bound has been derived, and an adversarial setting (e.g. \citet{rangi2018unifying, immorlica2019adversarial}), where the sublinear regret is impossible to obtain and a $O(\log T)$ competitive ratio has been derived. Similar to our DAL algorithm, the algorithms in the previous literature reply on an interplay between the primal and dual LPs. To be more concrete, \cite{agrawal2014bandits} and \cite{agrawal2014fast} develop a dual-based algorithms to BwK and a more general online stochastic optimization problem (belonging to bandits-based model) and analyzes the algorithm performance under further stronger conditions on the dual optimal solution. However, these learning models and algorithms are developed in the stationary (stochastic) environment, which cannot be applied to the non-stationary setting. For the non-stationary setting, the recent work \cite{liu2022non} derives sublinear regret, based on a more involved complexity measure over the non-stationarity of the underlying distributions which concerns both the temporary changes of two neighborhood distributions and the global changes of the entire distribution sequence.

Another representative problem for the bandits-based model is the OCO problem, which is one of the leading online learning frameworks \citep{hazan2016introduction}. Note that the standard OCO problem generally adopts a static optimal policy as the benchmark, i.e., the decision of the benchmark needs to be the same for each period. In contrast, in our model, the benchmark is a more powerful dynamic optimal policy where the decisions are allowed to be non-homogeneous across time. Therefore, not only our model is more involved, our benchmark is also stronger.
There have been results that consider a dynamic optimal policy as the benchmark for OCO \citep{besbes2015non, hall2013dynamical, jadbabaie2015online}, but all these works consider the unconstrained setting with no long-term constraints. For the line of works that study the problem of online convex optimization with constraints (OCOwC), existing literature would assume the constraint functions that characterize the long-term constraints are either static \citep{jenatton2016adaptive,yuan2018online,yi2021regret} or stochastically generated \citep{neely2017online}.

For the type-based model, one representative problem is the online packing problem, where the columns and the corresponding coefficient in the objective of the underlying LP come one by one and the decision has to be made on-the-fly. The packing problem covers a wide range of applications, including secretary problem \citep{ferguson1989solved, arlotto2019uniformly}, online knapsack problem \citep{arlotto2020logarithmic}, resource allocation problem \citep{li2020simple}, network routing problem \citep{buchbinder2009online}, matching problem \citep{mehta2007adwords}, capacity allocation problem \citep{zhong2018resource, lyu2019capacity, jiang2022achieving} etc. The problem is usually studied under either a stochastic model where the reward and size of each query is drawn independently from an unknown distribution $\mathcal{P}$, or a more general the random permutation model where the queries arrive in a random order  \citep{molinaro2014geometry, agrawal2014dynamic, kesselheim2014primal, gupta2014experts}. The more general online allocation problem (e.g. \cite{balseiro2022best}) has also been considered in the literature, where the objective and the constraint functions are allowed to be general functions.

\subsection{Road Map}
In \Cref{sec:formulation}, we introduce our formal problem formulation and state the assumptions we made in the paper. In \Cref{sec:stationary}, we consider a stationary setting where the demand distributions are homogeneous over time. We illustrate the main idea of our algorithmic design and present our results. We further investigate the development of our algorithm under the adversarial setting in \Cref{sec:adversarial}. Moreover, in \Cref{sec:nonstationary}, we explore the improvement of our algorithm and analysis with the existence of predictions. Numerical results are presented in \Cref{sec:numerical}. We present the proofs of our main theorems (\Cref{thm:RegretISP} and \Cref{thm:NStaRegretISP}) in the main body. All the other missing proofs and the extensions are presented in the supplementary material.


\section{Problem Formulation}\label{sec:formulation}
We consider the online two-stage stochastic optimization problem with long-term constraints in the following general formulation. There is a finite horizon of $T$ periods and at each period $t$, the following events happen in sequence:
\begin{itemize}
  \item[1] we decide the first-stage decision $\bm{c}_t\in\mathcal{C}$ and incur a cost $p(\bm{c}_t)$;
  \item[2] the type $\theta_t$ is drawn independently from an \textit{unknown} distribution $P_t$, and the second-stage objective function $f_{\theta_t}(\cdot)$ and the constraint function $\bm{g}_{\theta_t}(\cdot)=(g_{i,\theta_t}(\cdot))_{i=1}^m$ become known.
  \item[3] we decide the second-stage decision $\bm{x}_t\in\mathcal{K}(\theta_t, \bm{c}_t)$, where $\mathcal{K}(\theta_t, \bm{c}_t)$ is a feasible set parameterized by both the type $\theta_t$ and the first-stage decision $\bm{c}_t$, and incur an objective value $f_{\theta_t}(\cdot)$.
\end{itemize}
At the end of the entire horizon, the long-term constraints $\frac{1}{T}\cdot\sum_{t=1}^{T}\bx_t\in\mathcal{B}(\bm{C}, \bm{\theta})$ need to be satisfied, which is characterized as follows, following \cite{mahdavi2012trading},
\begin{equation}\label{eqn:LongConstraint}
  \frac{1}{T}\cdot\sum_{t=1}^{T}\bm{g}_{\theta_t}(\bc_t, \bm{x}_t)\leq\bm{\beta},
\end{equation}
with $\bm{\beta}\in(0,1)^m$. We aim to minimize the objective
\begin{equation}\label{eqn:LongObjective}
\sum_{t=1}^{T}\left(p(\bm{c}_t)+f_{\theta_t}(\bm{x}_t)\right).
\end{equation}
Any online policy is feasible as long as $\bm{c}_t$ and $\bm{x}_t$ are \textit{agnostic} to the future realizations while satisfying \eqref{eqn:LongConstraint}. The benchmark is the \textit{optimal policy}, denoted by $\pi^*$, who is aware of the distributions $\bm{P}=(P_t)_{t=1}^T$ but still the decisions $\bm{c}_t$ and $\bm{x}_t$ have to be agnostic to future realizations. Note that this benchmark is more power than the optimal online policy we are seeking for who is unaware of the distributions $\bm{P}$, and the optimal policy can be dynamic.
We are interested in developing a feasible online policy $\pi$ with known \textit{regret} upper bound compared to the optimal policy:
\begin{equation}\label{def:regret}
  \Reg(\pi, T)=\mathbb{E}_{\bm{\theta}\sim\bm{P}}\left[\ALG(\pi,\bm{\theta})-\ALG(\pi^*, \bm{\theta})\right],
\end{equation}
where $\ALG(\pi,\bm{\theta})$ denotes the objective value of policy $\pi$ on the sequence $\bm{\theta}$. The optimal policy can possess very complicated structures thus lacks tractability. Therefore, we develop a tractable lower bound of $\mathbb{E}_{\bm{\theta}\sim\bm{P}}[\ALG(\pi^*, \bm{\theta})]$. The lower bound is given by the optimization problem below.
\begin{align}
 \tag{OPT}\OPT= \min & ~~\sum_{t=1}^{T}\mathbb{E}_{\tbc_t, \tbx_t, \theta_t}\left[p(\tilde{\bm{c}}_t)+f_{\theta_t}(\tilde{\bm{x}}_t)\right], \label{lp:OPT}\\
  \mbox{s.t.} & ~~\frac{1}{T}\cdot\sum_{t=1}^{T}\mathbb{E}_{\tbc_t, \tbx_t, \theta_t}[\bm{g}_{\theta_t}(\tbc_t, \tbx_t)]\leq\bm{\beta}, \nonumber\\
  & ~~\tbx_t\in\mathcal{K}(\theta_t, \tbc_t), \tbc_t\in\mathcal{C}, \forall t.\nonumber
\end{align}
Here, $\tbc_t$ and $\tbx_t$ are random variables for each $t\in[T]$ and the distribution of $\tbc_t$ is \textit{independent} of $\theta_t$. Note that we only need the formulation of \eqref{lp:OPT} to conduct our theoretical analysis. We never require to really solve the optimization problem \eqref{lp:OPT}.
Clearly, if we let $\tbc_t$ (resp. $\tbx_t$) denote the \textit{marginal distribution} of the first-stage (resp. second-stage) decision made by the optimal policy, the we would have a feasible solution to \Cref{lp:OPT} while the objective value equals $\mathbb{E}_{\bm{\theta}\sim\bm{P}}[\ALG(\pi^*, \bm{\theta})]$. This argument leads to the following lemma.
\begin{lemma}[forklore]\label{lem:LowerBound}
 $\OPT\leq \mathbb{E}_{\bm{\theta}\sim\bm{P}}[\ALG(\pi^*, \bm{\theta})]$.
\end{lemma}
Throughout the paper, we make the following assumptions.
\begin{assumption}\label{assump:main}
The following conditions are satisfied:
\begin{itemize}
  \item[a.] (convexity) The function $p(\cdot)$ is a convex function with $p(\bm{0})=0$ and $\mathcal{C}$ is a compact and convex set which contains $\bm{0}$. The set $\mathcal{A}\subset[0,1]^m$ is a convex set. Also, the functions $f_{\theta}(\cdot)$ is convex in $\bm{x}$ and $\bm{g}_{\theta}(\cdot, \cdot)$ is convex in both $\bc$ and $\bm{x}$, for any $\theta$.
  \item[b.] (compactness) The set $\mathcal{K}(\theta_t, \bm{c}_t)$ is a polyhedron given by $\mathcal{K}(\theta_t, \bm{c}_t)=\{\bm{x}\in\mathbb{R}^m: \bm{0}\leq\bm{x}\text{~and~}B_{\theta_t}\bm{x}\leq\bm{c}_t\}\cap\mathcal{K}$ where $\mathcal{K}$ is a compact convex set.
  \item[c.] (boundedness) For any $\bm{x}\in\mathcal{K}$ and any $\theta$, we have $f_{\theta}(\bm{x})\in[-1,1]$ and $\bm{g}_{\theta}(\bm{x})\in[0,1]^m$. Moreover, it holds that $f_{\theta}(\bm{0})=0$ and $\bm{g}_{\theta}(\bm{0}, \bm{0})=\bm{0}$.
\end{itemize}
\end{assumption}

The conditions listed in \Cref{assump:main} are standard in the literature, which mainly ensure the convexity of the problem as well as the boundedness of the objective functions and the feasible set. Finally, in condition c, we assume that there always exists $\bm{0}$, which denotes a \textit{null} action that has no influence on the accumulated objective value and the constraints.

The Lagrangian dual problem of \eqref{lp:OPT} is given below:
\begin{align}
\tag{Dual}\max_{\bm{\lambda}\geq0}\min_{\tbc_t\in\mathcal{C}}~~ \LInner(\bm{C}, \bm{\lambda})=\sum_{t=1}^{T}\mathbb{E}\left[p(\tbc_t)-\frac{1}{T}\cdot\sum_{i=1}^{m}\lambda_i+
 \min_{\tbx_t\in\mathcal{K}(\theta_t,\tbc_t)}f_{\theta_t}(\tbx_t)+\sum_{i=1}^{m}\frac{\lambda_i\cdot g_{i,\theta_t}(\tbc_t, \tbx_t)}{T\cdot\beta_i}\right].  \label{lp:DualInner}
\end{align}
where $\bm{C}=(\tbc_t)_{t=1}^T$ and we note that the distribution of $\tbc_t$ is independent of $\theta_t$.
From weak duality, \eqref{lp:DualInner} also serves as a lower bound of $\mathbb{E}_{\bm{\theta}\sim\bm{P}}[\ALG(\pi^*, \bm{\theta})]$.

\section{Online Algorithm for Stationary Setting}\label{sec:stationary}
In this section, we consider the stationary setting where $P_t=P$ for each $t\in[T]$. Before formally presenting our algorithm and the corresponding regret analysis, we provide intuitions on the algorithmic design, based on relating the minimax dual problem \eqref{lp:DualInner} to the \textit{zero-sum game} that is described below. The connection between the primal-dual problem and the zero-sum game has been explored in \cite{roth2016watch, rivera2018online, immorlica2019adversarial} for bandit optimization, and we further develop this idea for our online two-stage problem.

\noindent\textbf{Minimizing regret in repeated zero-sum games.} Our algorithm is built from the \textit{zero-sum game}, which is a game between two players $j\in\{1,2\}$ with action sets $A_1$ and $A_2$ and payoff matrix $H\in\mathbb{R}^{|A_1|\times|A_2|}$. At each period, player 1 choose an action $a_1\in A_1$ and player 2 choose an action $a_2\in A_2$ and the outcome is $H_{a_1,a_2}$. Player 1 receives $H_{a_1,a_2}$ as \textit{reward} while player 2 receives $H_{a_1,a_2}$ as \textit{cost}. There is an algorithm $\ALG_1$ for player 1 to maximize the reward over the entire horizon and there is an algorithm $\ALG_2$ for player 2 to minimize the total cost. The game is \textit{stochastic} if the outcome $H_{a_1,a_2}$ is drawn independently from a given distribution that depends on $a_1$ and $a_2$.

The minimax Lagrangian dual problem \eqref{lp:DualInner} defines a zero-sum game. To see this point, we first note that under the stationary setting, it is optimal to set an identical first-stage decision, i.e. $\bm{c}^*_t=\bm{c}^*$ for some $\bm{c}^*$. Then, we define
\begin{align}
\bLInner(\bm{c}, \bm{\lambda}, \theta)=&p(\bm{c})-\frac{1}{T}\cdot\sum_{i=1}^{m}\lambda_i+\min_{\bm{x}\in\mathcal{K}(\theta,\bm{c})}\left\{f_{\theta}(\bm{x})+\sum_{i=1}^{m}\frac{\lambda_i\cdot g_{i,\theta}(\bc, \bm{x})}{T\cdot\beta_i}\right\},  \label{eqn:LISPstationary}
\end{align}
as the single-period decomposition of $\LInner(\bm{c},\bm{\lambda})$ with realization $\theta$.
\begin{lemma}\label{lem:StaDual}
Under the stationary setting where $P_t=P$ for each $t\in[T]$, it holds that
\begin{equation}\label{eqn:101001}
\max_{\bm{\lambda}\geq0}\min_{\tbc_t\in\mathcal{C}} \mathbb{E}_{\tilde{\bm{C}}}[\LInner(\bm{C}, \bm{\lambda})]=\max_{\bm{\lambda}\geq0}\min_{\tbc\in\mathcal{C}} T\cdot\mathbb{E}_{\tbc, \theta\sim P}\left[\bLInner(\tbc, \bm{\lambda}, \theta)\right].
\end{equation}
\end{lemma}
The zero-sum game can be described as follows. At each period $t$, player 1 chooses the first-stage decision $\bm{c}_t\in\mathcal{C}$, and play 2 chooses the dual variable $\bm{\lambda}_t\geq0$. We regard $\bm{\lambda}_t$ as a distribution over the long-term constraints, after divided by a scaling factor $\mu$. Then, the range of the dual variable for all the minimax problems \eqref{eqn:101001} can be restricted to the set $\mu\cdot \Delta_m$ where $\Delta_m=\{\bm{y}\in\mathbb{R}_{\geq0}^m:\sum_{i=1}^{m}y_i=1 \}$ denotes a distribution over the long-term constraints and $\mu>0$ is a constant to be specified later. The player 2 actually chooses one constraint $i_t$ among the long-term constraints, by setting $\bm{\lambda}_t=\mu\cdot\bm{e}_{i_t}$ where $\bm{e}_{i_t}\in\mathbb{R}^m$ is a vector with 1 as the $i_t$-th component and 0 for all other components. Clearly, what we can observe is a \textit{stochastic} outcome. For player 1, we can observe the stochastic outcome $\bLInner(\bm{c}_t,\mu\cdot\bm{e}_{i_t},\theta_t)$, where $\theta_t$ is one sample drawn independently from the distribution $P$. For player 2, we can observe the stochastic outcome $\bLInner(\bm{c}_t,\mu\cdot\bm{e}_{i_t},\theta_t)$ for the currently chosen constraint $i_t$. Finally, the second-stage decision $\bm{x}_t$ is determined by solving the inner minimization problem in \eqref{eqn:LISPstationary} for $\bm{c}=\bm{c}_t$ and $\bm{\lambda}=\mu\cdot\bm{e}_{i_t}$. Here, $\bm{c}_t$ is determined by $\ALG_1$ for player 1 and $i_t$ is determined by $\ALG_2$ for player 2.

We now further specify what is the observed outcome for player 2. Note that player 2 choose a constraint $i_t$ as action. The corresponding outcome is $\bLInner_{i_t}(\bm{c}_t, \mu\cdot\bm{e}_{i_t},\theta_t)$, where $\bLInner_{i}(\bm{c}_t, \mu\cdot\bm{e}_{i_t},\theta_t)$ is defined as follows for each $i\in[m]$,
\begin{equation}\label{eqn:defOutISP}
\bLInner_{i}(\bm{c}_t, \mu\cdot\bm{e}_{i_t},\theta_t)=p(\bm{c}_t)-\frac{\mu}{T}+f_{\theta_t}(\bm{x}_t)+\frac{\mu\cdot g_{i,\theta_t}(\bc_t, \bm{x}_t)}{T\cdot\beta_i},
\end{equation}
where $\bm{x}_t$ denotes the second-stage decision we made at period $t$.
Though the action for player 2 is $\bm{e}_{i_t}$, we are actually able to obtain \textit{additional information} for player 2, which helps the convergence of $\ALG_2$. It is easy to see that we have the value of $\bLInner_i(\bm{c}_t,\mu\cdot\bm{e}_{i_t},\theta_t)$ at the end of period $t$, for all other constraint $i\neq i_t$. Therefore, for $\ALG_2$, we are having the \textit{full feedback}, where the outcomes of all constraints, $\bLInner_i$ for all $i\in[m]$, can be observed. It also worth noting that under \Cref{assump:main}, $\bLInner(\bm{c}, \bm{\lambda}, \theta)$ is a convex functions over $\bm{c}$, for any fixed $\bm{\lambda}$ and $\theta$.
\begin{lemma}\label{lem:ConvexL}
For any $\bm{\lambda}$ and any $\theta$, $\bLInner(\bm{c}, \bm{\lambda}, \theta)$ defined in \eqref{eqn:LISPstationary} is a convex function over $\bm{c}$, under \Cref{assump:main}.
\end{lemma}
Following the above discussion, in practice, we can select \textit{Online Gradient Descent} (OGD) algorithm \citep{zinkevich2003online} as $\ALG_1$ for player 1, and we can select \textit{Hedge} algorithm \citep{freund1997decision} as $\ALG_2$ for player 2, where we regard $m$ long-term constraints as $m$ experts. Both algorithms work for adversarial input, which corresponds to the \textit{non-homogeneity} of the input $\bm{\lambda}_t$ of $\ALG_1$ for player 1 and the input $\bm{c}_t$ of $\ALG_2$ for player $2$. Therefore, we name our algorithm \textit{Doubly Adversarial Learning} (DAL) algorithm. The formal algorithm is given in \Cref{alg:DAL}.

\begin{algorithm}[tb]
\caption{Doubly Adversarial Learning (DAL) algorithm}
\label{alg:DAL}
\begin{algorithmic}
\STATE {\bfseries Input:} the scaling factor $\mu>0$, the adversarial learning algorithm $\ALG_1$ for player 1, the adversarial learning algorithm $\ALG_2$ for player 2.
\FOR{$t=1,\dots,T$}
\STATE $\bm{1}.$ $\ALG_1$ returns$\bm{c}_t$ and $\ALG_2$ returns $i_t\in[m]$.
\STATE $\bm{2}.$ Observe $\theta_t$ and determine $\bm{x}_t$ by solving the inner problem in \eqref{eqn:LISPstationary} with $\bm{c}=\bm{c}_t$ and $\bm{\lambda}=\mu\cdot\bm{e}_{i_t}$.
\STATE $\bm{3}.$ Return $\bLInner(\bm{c}_t,\mu\cdot\bm{e}_{i_t},\theta_t)$ to $\ALG_1$.
\STATE $\bm{4}.$ Return $\bLInner_i(\bm{c}_t,\mu\cdot\bm{e}_{i_t},\theta_t)$ defined in \eqref{eqn:defOutISP} for all $i\in[m]$ to $\ALG_2$.
\IF{$\frac{1}{T}\cdot\sum_{\tau=1}^{t}g_{i,\theta_{\tau}}(\bc_{\tau}, \bm{x}_{\tau})>\beta_i$ for some $i\in[m]$}
\STATE we terminate the algorithm by taking the null action $\bm{0}$ for both stage decision in the remaining horizon.
\ENDIF
\ENDFOR
\end{algorithmic}
\end{algorithm}
We denote by $\text{Reg}_1(T, \bm{\theta})$ the regret of $\ALG_1$ for player 1, given the sequence of realizations $\bm{\theta}$. In our setting, $\text{Reg}_1(T, \bm{\theta})$ will be the standard regret bound for OGD algorithm, which we formalize in \Cref{thm:OGDregret}. Similarly, we denote by $\text{Reg}_2(T, \bm{\theta})$ the regret of $\ALG_2$ for player 2, given the sequence of realizations $\bm{\theta}$. In our setting, $\text{Reg}_2(T, \bm{\theta})$ will be the standard regret bound for Hedge algorithm, which we formalize in \Cref{thm:Hedgeregret}. The regret of \Cref{alg:DAL} can be bounded by $\text{Reg}_1(T, \bm{\theta})$ and $\text{Reg}_2(T, \bm{\theta})$, as shown in the following theorem.
\begin{theorem}\label{thm:RegretISP}
Denote by $\pi$ \Cref{alg:DAL} with input $\mu=T$. Then, under \Cref{assump:main}, the regret enjoys the upper bound
\begin{equation}\label{eqn:121301}
\text{Regret}(\pi, T)\leq \mathbb{E}_{\bm{\theta}\sim\bm{P}}\left[\text{Reg}_1(T, \bm{\theta})+ \text{Reg}_2(T, \bm{\theta}) \right].
\end{equation}
Moreover, if OGD is selected as $\ALG_1$ and Hedge is selected as $\ALG_2$, the regret enjoys the upper bound
\begin{equation}\label{eqn:121302}
\text{Regret}(\pi, T)\leq \tilde{O}((G+F)\sqrt{T})+\tilde{O}(\sqrt{T\cdot\log m}),
\end{equation}
where $F$ is an upper bound of the diameter of the set $\mathcal{C}$ and $G$ is a constant that depends only on (the upper bound of gradients of) $\{f_{\theta}, \bm{g}_{\theta}\}_{\forall \theta}$, the minimum positive element of $B_{\theta}$ for all $\theta$, and $\max_{i\in[m]}\{\frac{1}{\beta_i}\}$.
\end{theorem}
\begin{myproof}[Proof of \Cref{thm:RegretISP}]
Denote by $\tau$ the time period that \Cref{alg:DAL} is terminated. There must be a constraint $i'\in[m]$ such that
\begin{equation}\label{eqn:121402}
\sum_{t=1}^{\tau} g_{i', \theta_t}(\bc_t, \bm{x}_t)\geq T\cdot \beta_{i'}.
\end{equation}
Otherwise, we can assume without loss of generality that there exists a \textit{dummy} constraint $i'$ such that $g_{i',\theta}(\bc, \bm{x})=\beta_{i'}=\alpha$, for arbitrary $\alpha\in(0,1)$, for any $\theta$ and $\bc, \bm{x}$. In this case, we can set $\tau=T$.

We denote by $(\bm{\lambda}^*, \tbc^*)$ (note $\tbc^*$ is allowed to be random) one \textit{saddle-point} optimal solution to
\[
\max_{\bm{\lambda}\geq0}\min_{\tbc\in\mathcal{C}} \mathbb{E}_{\tbc, \theta\sim P}\left[\bLInner(\tbc, \bm{\lambda}, \theta)\right]=\min_{\tbc\in\mathcal{C}}\max_{\bm{\lambda}\geq0} \mathbb{E}_{\tbc, \theta\sim P}\left[\bLInner(\tbc, \bm{\lambda}, \theta)\right].
\]
We have for any $\bc$ that
\[
\sum_{t=1}^{\tau}\bLInner(\bm{c}_t, \mu\cdot\bm{e}_{i_t}, \theta_t)\leq \sum_{t=1}^{\tau}\bar{L}(\bm{c}, \mu\cdot\bm{e}_{i_t}, \theta_t)+\text{Reg}_1(\tau, \bm{\theta})
\]
following regret bound of $\ALG_1$, which implies that
\[
\sum_{t=1}^{\tau}\bLInner(\bm{c}_t, \mu\cdot\bm{e}_{i_t}, \theta_t)\leq \sum_{t=1}^{\tau}\mathbb{E}_{\tbc^*}\left[\bar{L}(\tbc^*, \mu\cdot\bm{e}_{i_t}, \theta_t)\right]+\text{Reg}_1(\tau, \bm{\theta}).
\]
Then, it holds that
\begin{equation}\label{eqn:121401}
\begin{aligned}
\mathbb{E}_{\bm{\theta}\sim\bm{P}}\left[ \sum_{t=1}^{\tau}\bLInner(\bm{c}_t, \mu\cdot\bm{e}_{i_t}, \theta_t) \right]&\leq \mathbb{E}_{\tbc^*, \bm{\theta}\sim\bm{P}}\left[ \sum_{t=1}^{\tau}\bLInner(\tbc^*, \mu\cdot\bm{e}_{i_t}, \theta_t) \right]+\mathbb{E}_{\bm{\theta}\sim\bm{P}}\left[\text{Reg}_1(\tau, \bm{\theta})\right]\\
&\leq \tau\cdot \mathbb{E}_{\tbc^*, \theta\sim P}\left[ \bLInner(\tbc^*, \bm{\lambda}^*, \theta) \right]+\mathbb{E}_{\bm{\theta}\sim\bm{P}}\left[\text{Reg}_1(\tau, \bm{\theta})\right].
\end{aligned}
\end{equation}
where the last inequality follows from the definition of the saddle-point $(\bm{\lambda}^*, \tbc^*)$.
On the other hand, for any $i\in[m]$, we have
\[
\sum_{t=1}^{\tau}\bLInner(\bm{c}_t, \mu\cdot\bm{e}_{i_t}, \theta_t)\geq \sum_{t=1}^{\tau}\bLInner_i(\bm{c}_t, \mu\cdot\bm{e}_{i_t}, \theta_t)-\text{Reg}_2(\tau, \bm{\theta})
\]
following the regret bound of $\ALG_2$ (holds for arbitrary $\bm{\lambda}=\mu\cdot\bm{e}_i$). We now set $i=i'$ and we have
\[\begin{aligned}
\sum_{t=1}^{\tau}\bLInner(\bm{c}_t, \mu\cdot\bm{e}_{i_t}, \theta_t)&\geq \sum_{t=1}^{\tau}\bLInner_{i'}(\bm{c}_t, \mu\cdot\bm{e}_{i_t}, \theta_t)-\text{Reg}_2(\tau, \bm{\theta})\\
&=\sum_{t=1}^{\tau} \left(p(\bm{c}_t)+f_{\theta_t}(\bm{x}_t)\right)-\mu\cdot \frac{\tau}{T}+\sum_{t=1}^{\tau}\frac{\mu\cdot g_{i',\theta_t}(\bc_t, \bm{x}_t)}{T\cdot\beta_{i'}}-\text{Reg}_2(\tau, \bm{\theta})\\
&\geq \sum_{t=1}^{\tau} \left(p(\bm{c}_t)+f_{\theta_t}(\bm{x}_t)\right)+\mu\cdot\frac{T-\tau}{T}-\text{Reg}_2(\tau, \bm{\theta})
\end{aligned}\]
where the last inequality follows from \eqref{eqn:121402}. Then, it holds that
\begin{equation}\label{eqn:121403}
\mathbb{E}_{\bm{\theta}\sim\bm{P}}\left[ \sum_{t=1}^{\tau}\bLInner(\bm{c}_t, \mu\cdot\bm{e}_{i_t}, \theta_t) \right]\geq \mathbb{E}_{\bm{\theta}\sim\bm{P}}\left[ \sum_{t=1}^{\tau} \left(p(\bm{c}_t)+f_{\theta_t}(\bm{x}_t)\right) \right]+\mu\cdot\frac{T-\tau}{T}-\mathbb{E}_{\bm{\theta}\sim\bm{P}}\left[ \text{Reg}_2(\tau, \bm{\theta}) \right].
\end{equation}
Combining \eqref{eqn:121401} and \eqref{eqn:121403}, we have
\[
\mathbb{E}_{\bm{\theta}\sim\bm{P}}\left[ \sum_{t=1}^{\tau} \left(p(\bm{c}_t)+f_{\theta_t}(\bm{x}_t)\right) \right]\leq -\mu\cdot\frac{T-\tau}{T}+\tau\cdot \mathbb{E}_{\theta\sim P}\left[ \bLInner(\tbc^*, \bm{\lambda}^*, \theta) \right]+\mathbb{E}_{\bm{\theta}\sim\bm{P}}\left[\text{Reg}_1(\tau, \bm{\theta})\right]+\mathbb{E}_{\bm{\theta}\sim\bm{P}}\left[ \text{Reg}_2(\tau, \bm{\theta}) \right].
\]
From the boundedness conditions in \Cref{assump:main}, we have
\[
\mathbb{E}_{\tbc^*, \theta\sim P}\left[\bLInner(\tbc^*, \bm{\lambda}^*, \theta)\right]=\frac{1}{T}\cdot\OPTInner\geq -1
\]
which implies that
\[
-\mu\cdot\frac{T-\tau}{T}\leq (T-\tau)\cdot \mathbb{E}_{\tbc^*, \theta\sim P}\left[\bLInner(\tbc^*, \bm{\lambda}^*, \theta)\right]
\]
when $\mu=T$. Therefore, we have
\begin{equation}\label{eqn:121404}
\begin{aligned}
\mathbb{E}_{\bm{\theta}\sim\bm{P}}\left[ \sum_{t=1}^{\tau} \left(p(\bm{c}_t)+f_{\theta_t}(\bm{x}_t)\right) \right]&\leq T\cdot \mathbb{E}_{\tbc^*, \theta\sim P}\left[ \bLInner(\bm{c}^*, \bm{\lambda}^*, \theta) \right]+\mathbb{E}_{\bm{\theta}\sim\bm{P}}\left[\text{Reg}_1(\tau, \bm{\theta})\right]+\mathbb{E}_{\bm{\theta}\sim\bm{P}}\left[ \text{Reg}_2(\tau, \bm{\theta}) \right]\\
&\leq T\cdot \mathbb{E}_{\tbc^*, \theta\sim P}\left[ \bLInner(\tbc^*, \bm{\lambda}^*, \theta) \right]+\mathbb{E}_{\bm{\theta}\sim\bm{P}}\left[\text{Reg}_1(T, \bm{\theta})\right]+\mathbb{E}_{\bm{\theta}\sim\bm{P}}\left[ \text{Reg}_2(T, \bm{\theta}) \right].
\end{aligned}
\end{equation}
which completes our proof of \eqref{eqn:121301}. Using \Cref{thm:OGDregret} and \Cref{thm:Hedgeregret} to bound $\mathbb{E}_{\bm{\theta}\sim\bm{P}}\left[\text{Reg}_1(T, \bm{\theta})\right]$ and $\mathbb{E}_{\bm{\theta}\sim\bm{P}}\left[ \text{Reg}_2(T, \bm{\theta}) \right]$, we have
\[
\mathbb{E}_{\bm{\theta}\sim\bm{P}}\left[ \sum_{t=1}^{\tau} \left(p(\bm{c}_t)+f_{\theta_t}(\bm{x}_t)\right) \right]\leq T\cdot \mathbb{E}_{\tbc^*, \theta\sim P}\left[ \bLInner(\tbc^*, \bm{\lambda}^*, \theta) \right]+O((G+F)\cdot\sqrt{T})+O(\sqrt{T\cdot\log m})
\]
which completes our proof of \eqref{eqn:121302}.
\end{myproof}

Notably, when the first-stage decision $\bc_t$ is de-activated for each period $t$, i.e., $\mathcal{C}$ is a singleton, then there is no need to run $\ALG_1$ to update $\bc_t$ and our problem would reduce to the online allocation problem studied in \cite{balseiro2022best} and \cite{jiang2020online}. The regret upper bound $\text{Regret}(\pi, T)$ in \Cref{thm:RegretISP} would reduce to $\mathbb{E}_{\bm{\theta}\sim\bm{P}}[\text{Reg}_2(T,\bm{\theta})]$ and thus $\tilde{O}(\sqrt{T\cdot\log m})$ if Hedge is selected as $\ALG_2$. In terms of the dependency on $m$, our regret bound improves upon the $O(\sqrt{mT})$ regret in \cite{balseiro2022best} where a mirror descent algorithm is used to update the dual variable $\bm{\lambda}_t$ and the $O(m\sqrt{T})$ regret in \cite{jiang2020online} where a stochastic gradient descent algorithm is used to update $\bm{\lambda}_t$. This improvement reveals the benefit of our \Cref{alg:DAL} that builds on a adversarial learning framework.

\section{Robustness to Adversarial Corruptions}\label{sec:adversarial}
In this section, we consider our problem with the presence of adversarial corruptions. To be specific, we assume that at each period $t$, after the type $\theta_t$ is realized, there can be an adversary corrupting $\theta_t$ into $\theta_t^c$, and only the value of $\theta^c_t$ is revealed to us. The adversarial corruption to a stochastic model can arise from the non-stationarity of the underlying distributions $\bm{P}$ \citep{jiang2020online}, or malicious attack and false information input to the system \citep{lykouris2018stochastic}. We now show that our \Cref{alg:DAL} can tolerate a certain amount of adversarial corruptions and the performance guarantees would degrade smoothly as the total number of corruptions increase from $0$ to $T$.

We first characterize the difficulty of the problem. Denote by $W(\bm{\theta})$ the total number of adversarial corruptions on sequence $\bm{\theta}$, i.e.,
\begin{equation}\label{eqn:Advercorrup}
  W(\bm{\theta})=\sum_{t=1}^{T}\bI(\theta_t\neq \theta_t^c).
\end{equation}
We show that the optimal regret bound scales at least $\Omega(\mathbb{E}_{\bm{\theta}\sim\bm{P}}[W(\bm{\theta})])$. Note that in the definition of the regret, the performance of our algorithm is the total collected reward \textit{after} adversarial corrupted, and the benchmark is the optimal policy \textit{with} adversarial corruptions, denoted by $\pi^*$.
\begin{theorem}\label{thm:adverLower}
Let $W(\bm{\theta})$ be the total number of adversarial corruptions on sequence $\bm{\theta}$, as defined in \eqref{eqn:Advercorrup}. For any feasible online policy $\pi$, there always there exists distributions $\bm{P}$ and a way to corrupt $\bm{P}$ such that
\begin{align}
\Reg^c(\pi, T)&=\mathbb{E}_{\bm{\theta}\sim\bm{P}}[\ALG(\pi,\bm{\theta}^c)]-\mathbb{E}_{\bm{\theta}\sim\bm{P}}[\ALG(\pi^*,\bm{\theta}^c)]\geq
\Omega\left(\mathbb{E}_{\bm{\theta}\sim\bm{P}}[W(\bm{\theta})]\right). \nonumber
\end{align}
\end{theorem}
Our lower bound in \Cref{thm:adverLower} is in correspondence to the lower bound established in \cite{lykouris2018stochastic} for stochastic multi-arm-bandits model. We now show that our \Cref{alg:DAL} achieves a regret bound that matches the lower bound in \Cref{thm:adverLower} in terms of the dependency on $W(\bm{\theta})$, which is in correspondence to the linear dependency on $W(\bm{\theta})$ established in \cite{gupta2019better} for the stochastic multi-arm-bandits model. Note that in the implementation of \Cref{alg:DAL}, $\theta_t$ is always replaced by $\theta^c_t$ which is the only type information that is revealed to us.

\begin{theorem}\label{thm:CorruptRegretISP}
Denote by $\pi$ \Cref{alg:DAL} with input $\mu=T$. Then, under \Cref{assump:main} and the corrupted setting, the regret enjoys the upper bound
\begin{align}
\text{Regret}^{\text{c}}(\pi, T)=\mathbb{E}_{\bm{\theta}\sim\bm{P}}[\ALG(\pi,\bm{\theta}^c)]-\mathbb{E}_{\bm{\theta}\sim\bm{P}}[\ALG(\pi^*,\bm{\theta}^c)] \leq &\tilde{O}((G+F)\cdot\sqrt{T})+\tilde{O}(\sqrt{T\cdot\log m})\nonumber\\
&+O\left( \mathbb{E}_{\bm{\theta}\sim\bm{P}}[W(\bm{\theta})] \right) , \label{eqn:121801}
\end{align}
if OGD is selected as $\ALG_1$ and Hedge is selected as $\ALG_2$. Here, $\mathbb{E}_{\bm{\theta}\sim\bm{P}}[W(\bm{\theta})]$ denotes the expectation of the total number of corruptions, with $W(\bm{\theta})$ defined in \eqref{eqn:Advercorrup}.
\end{theorem}

We remark that the implementation of our \Cref{alg:DAL} in this setting is \textit{agnostic} to the total number of adversarial corruptions and achieves the optimal dependency on the number of corruptions. This is one fascinating benefits of adopting adversarial learning algorithms as algorithmic subroutines in \Cref{alg:DAL}, where the corruptions can also be incorporated as adversarial input which is handled by the learning algorithms. On the other hand, even if the number of corruptions is given to us as a prior knowledge, \Cref{thm:adverLower} shows that still, no online policy can achieve a better dependency on the number of corruptions.

\section{Improvement of Adversarial Setting with Predictions}\label{sec:nonstationary}

In this section, we consider the adversarial setting where $P_t$ is \textit{unknown} and \textit{non-homogeneous} for each $t$. The adversarial setting can be captured by the setting in \Cref{sec:adversarial} if we allow an \textit{arbitrary} number of corruptions to stationary distributions. However, following \Cref{thm:adverLower}, we would inevitably have a linear dependency on the number of corruptions, which translates into a linear regret for the adversarial setting. Therefore, we now explore whether we can get an improved bound, with the help of \textit{additional information}. We assume that we have a prediction of $P_t$, denoted by $\hat{P}_t$, for each $t\in[T]$. In practice, instead of letting $P_t$ be completely unknown, we usually have multiple historical samples of $P_t$ from which we can use some machine learning methods to form a prediction of $P_t$ (e.g. the empirical distribution over the samples). We explore how to utilize the prediction $\hat{P}_t$ for each $t\in[T]$ to obtain an improved bound.

We first derive the regret lower bound and show how the regret should depend on the possible inaccuracy of the predictions. Following \cite{jiang2020online}, we measure the inaccuracy of $\hat{P}_t$ by Wasserstein distance (we refer interested readers to Section 6.3 of \cite{jiang2020online} for benefits of using Wasserstein distance in online decision making), which is defined as follows
\begin{equation}\label{eqn:Wasserdist}
W(\hat{P}_t, P_t):=\inf_{Q\in\mathcal{F}(\hat{P}_t,P_t)}\int d(\theta, \theta')dQ(\theta, \theta'),
\end{equation}
where $d(\theta, \theta')=\|(f_{\theta}, \bm{g}_{\theta})-(f_{\theta'}, \bm{g}_{\theta'})\|_{\infty}$ and $\mathcal{F}(\hat{P}_t, P_t)$ denotes the set of all joint distributions for $(\theta, \theta')$ with marginal distributions $\hat{P}_t$ and $P_t$. In the following theorem, we show that one cannot break a linear dependency on the total inaccuracy of the prior estimates, by modifying the proof of \Cref{thm:adverLower}.
\begin{theorem}\label{thm:priorLower}
Let $W_T=\sum_{t=1}^{T}W(\hat{P}_t, P_t)$ be the total measure of inaccuracy, where $W(\hat{P}_t, P_t)$ is defined in \eqref{eqn:Wasserdist}. For any feasible online policy $\pi$ that only knows predictions, there always exists $\bm{P}$ and $\hat{\bm{P}}$ such that
\[
\Reg(\pi, T)=\mathbb{E}_{\bm{\theta}\sim\bm{P}}[\ALG(\pi,\bm{\theta})-\ALG(\pi^*,\bm{\theta})]\geq
\Omega\left(W_T\right),
\]
where $\pi^*$ denotes the optimal policy that knows true distributions $\bm{P}$.
\end{theorem}

We then derive our online algorithm with a regret that matches the lower bound established above. Note that in the settings without predictions (\Cref{sec:stationary} and \Cref{sec:adversarial}), we employ adversarial learning algorithm for $\bm{c}_t$ to ``learn'' a good first-stage decision. However, when we have a prediction, a more efficient way would be to ``greedily'' select $\bm{c}_t$ with the help of predictions. In order to describe our idea, we denote by $\{\hat{\bm{\lambda}}^*, \{\hbc^*_t\}_{\forall t\in[T]}\}$ one optimal solution to
\begin{equation}\label{eqn:121901}
\max_{\bm{\lambda}\geq0}\min_{\bm{c}_t\in\mathcal{C}} \hLInner(\bm{C}, \bm{\lambda})=\mathbb{E}_{\bm{\theta}\sim\hat{\bm{P}}}\left[\sum_{t=1}^{T}\left(p(\bc_t)-\frac{1}{T}\cdot\sum_{i=1}^{m}\lambda_i+
\min_{\bx_t\in\mathcal{K}(\theta_t,\bc_t)}f_{\theta_t}(\bx_t)+\sum_{i=1}^{m}\frac{\lambda_i\cdot g_{i,\theta_t}(\bc_t, \bx_t)}{T\cdot\beta_i}\right)\right]
\end{equation}
where $\hat{\bm{P}}=(\hat{P}_t)_{t=1}^T$. Note that the value of \eqref{eqn:121901} is equivalent to the value of \eqref{lp:DualInner} if $\hat{\bm{P}}=\bm{P}$. Then, we define for each $i\in[m]$,
\begin{equation}\label{eqn:121902}
\hat{\beta}_{i,t}:=\mathbb{E}_{\theta\sim\hat{P}_t}\left[ g_{i,\theta}(\hbc^*_t, \hbx^*_t(\theta)) \right],
\end{equation}
with
\[
\hbx^*_t(\theta)\in\text{argmin}_{\bx_t\in\mathcal{K}(\theta,\hbc^*_t)}f_{\theta}(\bx_t)+\sum_{i=1}^{m}\frac{\hat{\lambda}^*_i\cdot g_{i,\theta}(\hbc^*_t, \bx_t)}{T\cdot\beta_i}.
\]
Here, $\hat{\bm{\beta}}_t$ can be interpreted as the \textit{predictions-informed} target levels to achieve, at each period $t$. To be more concrete, if $\hat{P}_t=P_t$ for each $t$, then $\hat{\bm{\beta}}$ is exactly the value of the constraint functions at period $t$ in \eqref{lp:OPT}, which is a good reference to stick to. We then define
\begin{align}
&\hLInner_t(\bm{c}, \bm{\lambda}):=\mathbb{E}_{\theta\sim\hat{P}_t}\left[ p(\bc)-\sum_{i=1}^{m}\frac{\lambda_i\cdot\hat{\beta}_{i,t}}{T\cdot\beta_i}+\min_{\bx\in\mathcal{K}(\theta,\bc)}f_{\theta}(\bx)+\sum_{i=1}^{m}\frac{\lambda_i\cdot g_{i,\theta}(\bc, \bx)}{T\cdot\beta_i} \right]. \label{eqn:121903}
\end{align}
The key ingredient of our analysis is the following.
\begin{lemma}\label{lem:decompose}
It holds that
\begin{equation}\label{eqn:121904}
  \max_{\bm{\lambda}\geq0}\min_{\bm{c}_t\in\mathcal{C}} \hLInner(\bm{C}, \bm{\lambda})=\sum_{t=1}^{T}\max_{\bm{\lambda}\geq0}\min_{\bm{c}_t\in\mathcal{C}} \hLInner_t(\bm{c}_t, \bm{\lambda}),
\end{equation}
and moreover, letting $(\hat{\bm{\lambda}}^*, (\hbc^*_t)_{t=1}^T)$ be the optimal solution to $\max_{\bm{\lambda}\geq0}\min_{\bm{c}_t\in\mathcal{C}} \hLInner(\bm{C}, \bm{\lambda})$ used in the definition \eqref{eqn:121902}, then $(\hat{\bm{\lambda}}^*, \hbc^*_t)$ is an optimal solution to $\max_{\bm{\lambda}\geq0}\min_{\bm{c}_t\in\mathcal{C}} \hLInner_t(\bm{c}_t, \bm{\lambda})$ for each $t\in[T]$. Also, we have
\begin{equation}\label{eqn:121908}
\sum_{i=1}^{m}\sum_{t=1}^{T}\hat{\lambda}^*_{i}\cdot\hat{\beta}_{i,t}=T\cdot\sum_{i=1}^{m}\beta_i\cdot\hat{\lambda}^*_i,
\end{equation}
for each $i\in[m]$.
\end{lemma}
\Cref{lem:decompose} implies that given $\hat{\bm{\lambda}}^*$, we can simply minimize $\hLInner_t(\bm{c}_t, \hat{\bm{\lambda}}^*)$ over $\bm{c}_t$ to get the first-stage decision \footnote{This is a stochastic optimization problem, which can be solved by applying \textit{stochastic gradient descent} over $\bm{c}_t$ or applying \textit{sample average approximation} to get samples of $\theta\sim\hat{P}_t$}. As for $\bm{\lambda}$, we still apply adversarial learning algorithm (Hedge) to dynamically update it. Our formal algorithm is described in \Cref{alg:IAL}, which we call \textit{Informative Adversarial Learning} (IAL) algorithm since the updates are informed by the prior estimates.

\begin{algorithm}[tb]
\caption{Informative Adversarial Learning (IAL) algorithm}
\label{alg:IAL}
\begin{algorithmic}
\STATE {\bfseries Input:} the scaling factor $\mu>0$, the adversarial learning algorithm $\ALGD$ for dual variable $\bm{\lambda}$, and the prior estimates $\hat{\bm{P}}$.
\STATE {\bfseries Initialize:} compute $\hat{\beta}_{i,t}$ for all $i\in[m], t\in[T]$ as \eqref{eqn:121902}.
\FOR{$t=1,\dots,T$}
\STATE $\bm{1}.$ $\ALGD$ returns a long-term constraint $i_t\in[m]$.
\STATE $\bm{2}.$ Set $\bm{c}_t=\text{argmin}_{\bm{c}\in\mathcal{C}}\hLInner_t(\bm{c}, \mu\cdot\bm{e}_{i_t})$.
\STATE $\bm{3}.$ Observe $\theta_t$ and set $\bm{x}_t$ by solving the inner problem in \eqref{eqn:121903} with $\bm{c}=\bm{c}_t$ and $\bm{\lambda}=\mu\cdot\bm{e}_{i_t}$.
\STATE $\bm{4}.$ Return to $\ALGD$ $\hLInner_{i,t}(\bm{c}_t, \mu\cdot\bm{e}_{i_t}, \theta_t)$ defined as follows for each $i\in[m]$,
\begin{align}
\hLInner_{i,t}(\bm{c}_t, \mu\cdot\bm{e}_{i_t}, \theta_t)=&p(\bm{c}_t)-\frac{\mu\cdot\hat{\beta}_{i,t}}{T\cdot\beta_i}+f_{\theta_t}(\bm{x}_t) +\frac{\mu\cdot g_{i,\theta_t}(\bc_t, \bm{x}_t)}{T\cdot\beta_i}. \label{eqn:121916}
\end{align}
\ENDFOR
\end{algorithmic}
\end{algorithm}

As shown in the next theorem, the regret of \Cref{alg:IAL} is bounded by $O(W_T+\sqrt{T})$, which matches the lower bound established in \Cref{thm:priorLower}.

\begin{theorem}\label{thm:NStaRegretISP}
Denote by $\pi$ \Cref{alg:IAL} with input $\mu=\|\hat{\bm{\lambda}}^*\|_{\infty}=\alpha\cdot T$ for some constant $\alpha>0$. Denote by $W_T=\sum_{t=1}^{T}W(\hat{P}_t, P_t)$ the total measure of inaccuracy, with $W(\hat{P}_t, P_t)$ defined in \eqref{eqn:Wasserdist}. Then, under \Cref{assump:main}, the regret enjoys the upper bound
\begin{align}
\text{Regret}(\pi, T)=&\mathbb{E}_{\bm{\theta}\sim\bm{P}}[\ALG(\pi,\bm{\theta})]-\mathbb{E}_{\bm{\theta}\sim\bm{P}}[\ALG(\pi^*,\bm{\theta})]
\leq \tilde{O}(\sqrt{T\cdot\log m})+O( W_T )  \label{eqn:122001}
\end{align}
if Hedge is selected as $\ALGD$. Moreover, we have
\begin{align}
\frac{1}{T}\sum_{t=1}^{T}g_{i,\theta_t}(\bc_t, \bm{x}_t)-\beta_i\leq \tilde{O}\left(\sqrt{\frac{\log m}{T}}\right)+O\left(\frac{W_T}{ T}\right), \label{eqn:122002}
\end{align}
for each $i\in[m]$.
\end{theorem}
\begin{myproof}[Proof of \Cref{thm:NStaRegretISP}]
The proof can be classified by the following two steps. We denote by
\begin{equation}\label{eqn:121909}
\LInner_t(\bm{c}, \bm{\lambda}):=\mathbb{E}_{\theta\sim P_t}\left[ p(\bc)-\sum_{i=1}^{m}\frac{\lambda_i\cdot\hat{\beta}_{i,t}}{T\cdot\beta_i}+
\min_{\bx\in\mathcal{K}(\theta,\bc)}f_{\theta}(\bx)+\sum_{i=1}^{m}\frac{\lambda_i\cdot g_{i,\theta}(\bc, \bx)}{T\cdot\beta_i} \right]
\end{equation}
for each $t\in[T]$. Our first step is to show that for any $\bm{\lambda}\geq0$ and any $\bm{c}\in\mathcal{C}$, it holds that
\begin{equation}\label{eqn:121910}
  \left|\LInner_t(\bm{c}, \bm{\lambda})-\hLInner_t(\bm{c}, \bm{\lambda})\right|\leq \frac{\|\bm{\lambda}\|_1}{T\beta_{\min}}\cdot W(\hat{P}_t, P_t)
\end{equation}
where $\beta_{\min}=\min_{i\in[m]}\{\beta_i\}$.

We now prove \eqref{eqn:121910}. We define
\[
\hLInner_t(\bm{c}, \bm{\lambda}, \theta):= p(\bc)-\sum_{i=1}^{m}\frac{\lambda_i\cdot\hat{\beta}_{i,t}}{T\cdot\beta_i}+
\min_{\bx\in\mathcal{K}(\theta,\bc)}f_{\theta}(\bx)+\sum_{i=1}^{m}\frac{\lambda_i\cdot g_{i,\theta}(\bc, \bx)}{T\cdot\beta_i}.
\]
It is clear to see that
\[
\LInner_t(\bm{c}, \bm{\lambda})=\mathbb{E}_{\theta\sim P_t}\left[ \hLInner_t(\bm{c}, \bm{\lambda}, \theta) \right]\text{~and~}\hLInner_t(\bm{c}, \bm{\lambda})=\mathbb{E}_{\theta\sim \hat{P}_t}\left[ \hLInner_t(\bm{c}, \bm{\lambda}, \theta) \right].
\]
Moreover, note that for any $\theta, \theta'$, we have
\[
|\hLInner_t(\bm{c}, \bm{\lambda}, \theta)-\hLInner_t(\bm{c}, \bm{\lambda}, \theta')|\leq \frac{\|\bm{\lambda}\|_1}{T\beta_{\min}}\cdot d(\theta, \theta'),
\]
with $d(\theta, \theta')=\|(f_{\theta}, \bm{g}_{\theta})-(f_{\theta'}, \bm{g}_{\theta'})\|_{\infty}$ in the definition \eqref{eqn:Wasserdist}. Therefore, we have
\[
\left|\LInner_t(\bm{c}, \bm{\lambda})-\hLInner_t(\bm{c}, \bm{\lambda})\right|=\left| \mathbb{E}_{\theta\sim P_t}\left[\hLInner_t(\bm{c}, \bm{\lambda}, \theta)\right]-\mathbb{E}_{\theta'\sim\hat{P}_t}\left[\hLInner_t(\bm{c}, \bm{\lambda}, \theta')\right] \right|\leq \frac{\|\bm{\lambda}\|_1}{T\beta_{\min}}\cdot W(\hat{P}_t, P_t),
\]
thus, complete our proof of \eqref{eqn:121910}.

Our second step is to bound the final regret with the help of \eqref{eqn:121910}. We assume without loss of generality that there always exists $i'\in[m]$ such that
\begin{equation}\label{eqn:121915}
\sum_{t=1}^{T} g_{i', \theta_t}(\bc_t, \bm{x}_t)\geq T\cdot \beta_{i'}.
\end{equation}
In fact, let there be a \textit{dummy} constraint $i'$ such that $g_{i',\theta}(\bc, \bm{x})=\beta_{i'}=\alpha$, for arbitrary $\alpha\in(0,1)$, for any $\theta$ and $\bc, \bm{x}$. Then, \eqref{eqn:121915} holds.

Let $(\hat{\bm{\lambda}}^*, (\hbc^*_t)_{t=1}^T)$ be the optimal solution to $\max_{\bm{\lambda}\geq0}\min_{\bm{c}_t\in\mathcal{C}} \hLInner(\bm{C}, \bm{\lambda})$ used in the definition \eqref{eqn:121902}. Then, it holds that
\begin{align}
\mathbb{E}_{\bm{\theta}\sim\bm{P}}\left[ \sum_{t=1}^{T}\hLInner_t(\bm{c}_t, \mu\cdot\bm{e}_{i_t}, \theta_t) \right]&=\sum_{t=1}^{T}\mathbb{E}_{\bc_t, i_t}\left[\LInner_t(\bc_t, \mu\cdot\bm{e}_{i_t})\right]\label{eqn:121911}\\
&\leq\sum_{t=1}^{T}\mathbb{E}_{\bc_t, i_t}\left[\hLInner_t(\bm{c}_t, \mu\cdot\bm{e}_{i_t})\right]+\frac{\mu\cdot W_T}{T\beta_{\min}}=\sum_{t=1}^{T}\mathbb{E}_{i_t}\left[ \min_{\bc\in\mathcal{C}}\hLInner_t(\bm{c}, \mu\cdot\bm{e}_{i_t}) \right]+\frac{\mu\cdot W_T}{T\beta_{\min}}\nonumber\\
&\leq \sum_{t=1}^{T} \min_{\bc\in\mathcal{C}}\hLInner_t(\bm{c}, \hat{\bm{\lambda}}^*)+\frac{\mu\cdot W_T}{T\beta_{\min}}=\sum_{t=1}^{T} \hLInner_t(\hbc^*_t, \hat{\bm{\lambda}}^*)+\frac{\mu\cdot W_T}{T\beta_{\min}} \nonumber
\end{align}
where the first inequality follows from the definition of $\bc_t$, the second inequality follows from \Cref{lem:decompose}, and the last equality follows from the definition of $\hbc^*_t$.

On the other hand, for any $i\in[m]$, we have
\[
\sum_{t=1}^{T}\hLInner_t(\bm{c}_t, \mu\cdot\bm{e}_{i_t}, \theta_t)\geq \sum_{t=1}^{T}\hLInner_{i,t}(\bm{c}_t, \mu\cdot\bm{e}_{i_t}, \theta_t)-\text{Reg}(T, \bm{\theta})
\]
with $\hLInner_{i,t}$ defined in \eqref{eqn:121916}, where $\text{Reg}(\tau, \bm{\theta})$ denotes the regret bound of $\ALGD$ (holds for arbitrary $\bm{\lambda}=\mu\cdot\bm{e}_i$). We now denote by
\[
i^*=\text{argmax}_{i\in[m]}\{ \frac{1}{T}\cdot\sum_{t=1}^{T}g_{i,\theta_t}(\bc_t, \bm{x}_t)-\beta_i \}.
\]
We also denote by
\[
d_T(\mathcal{A}, \bm{\theta})=\max_{i\in[m]}\{ \frac{1}{T}\cdot\sum_{t=1}^{T}g_{i,\theta_t}(\bc_t, \bm{x}_t)-\beta_i \}.
\]
From \eqref{eqn:121915}, we must have $d_T(\mathcal{A}, \bm{\theta})\geq0$. We now set $i=i^*$ and we have
\begin{align}
\sum_{t=1}^{T}\hLInner_t(\bm{c}_t, \mu\cdot\bm{e}_{i_t}, \theta_t)&\geq \sum_{t=1}^{T}\hLInner_{i^*,t}(\bm{c}_t, \mu\cdot\bm{e}_{i_t}, \theta_t)-\text{Reg}(T, \bm{\theta}) \label{eqn:122201}\\
&=\sum_{t=1}^{T} \left(p(\bm{c}_t)+f_{\theta_t}(\bm{x}_t)\right)-\mu\cdot \frac{\sum_{t=1}^{T}\hat{\beta}_{i^*,t}}{T\beta_{i^*}}+\sum_{t=1}^{T}\frac{\mu\cdot g_{i^*,\theta_t}(\bc_t, \bm{x}_t)}{T\cdot\beta_{i^*}}-\text{Reg}(T, \bm{\theta}) \nonumber
\end{align}
From the construction of $\hat{\beta}_{i,t}$, we know that $\sum_{t=1}^{T}\hat{\beta}_{i,t}\leq T\cdot\beta_i$ for each $i\in[m]$. To see this point, we note that if we define $\hat{L}(\bm{\lambda})=\min_{\bm{c}_t\in\mathcal{C}} \hLInner(\bm{C}, \bm{\lambda})$, then, for each $i\in[m]$,
\begin{equation}\label{eqn:122202}
\nabla_{\lambda_i} \hat{L}(\hat{\bm{\lambda}}^*)=-1+\mathbb{E}_{\bm{\theta}\sim\hat{P}}\left[\sum_{t=1}^{T}\frac{g_{i,\theta_t}(\hbc^*_t, \hbx^*(\theta_t))}{T\beta_i}\right]
=-1+\frac{\sum_{t=1}^{T}\hat{\beta}_{i,t}}{T\cdot\beta_i}\leq 0.
\end{equation}
Otherwise, $\nabla_{\lambda_i} \hat{L}(\hat{\bm{\lambda}}^*)>0$ would violate the optimality of $\hat{\bm{\lambda}}^*$ to $\max_{\bm{\lambda}\geq0} \hat{L}(\bm{\lambda})$.

Plugging \eqref{eqn:122202} into \eqref{eqn:122201}, we get
\begin{align}
\sum_{t=1}^{T}\hLInner_t(\bm{c}_t, \mu\cdot\bm{e}_{i_t}, \theta_t)&\geq \sum_{t=1}^{T} \left(p(\bm{c}_t)+f_{\theta_t}(\bm{x}_t)\right)-\mu+\sum_{t=1}^{T}\frac{\mu\cdot g_{i^*,\theta_t}(\bc_t, \bm{x}_t)}{T\cdot\beta_{i^*}}-\text{Reg}(T, \bm{\theta})\label{eqn:121912}\\
&\geq \sum_{t=1}^{T} \left(p(\bm{c}_t)+f_{\theta_t}(\bm{x}_t)\right)+\frac{\mu}{\beta_{\min}}\cdot d_T(\mathcal{A}, \bm{\theta})-\text{Reg}(T, \bm{\theta})\nonumber
\end{align}
Combining \eqref{eqn:121911} and \eqref{eqn:121912}, we have
\begin{equation}\label{eqn:121918}
\mathbb{E}_{\bm{\theta}\sim\bm{P}}\left[ \sum_{t=1}^{T} \left(p(\bm{c}_t)+f_{\theta_t}(\bm{x}_t)\right) \right]\leq \sum_{t=1}^{T} \hLInner_t(\hbc^*_t, \hat{\bm{\lambda}}^*)+\frac{\mu\cdot W_T}{T\beta_{\min}}+\mathbb{E}_{\bm{\theta}\sim\bm{P}}\left[\text{Reg}(T, \bm{\theta})\right]-\frac{\mu}{\beta_{\min}}\cdot\mathbb{E}_{\bm{\theta}\sim\bm{P}}\left[ d_T(\mathcal{A}, \bm{\theta}) \right].
\end{equation}
Denote by $(\bm{\lambda}^*, \tilde{\bm{C}}^*)$ the optimal \textit{saddle-point} solution to
\[
\max_{\bm{\lambda}\geq0} \min_{\tilde{\bm{C}}\in\mathcal{C}} \mathbb{E}_{\tilde{\bm{C}}}[\LInner(\tilde{\bm{C}}, \bm{\lambda})]
=\min_{\tilde{\bm{C}}\in\mathcal{C}}\max_{\bm{\lambda}\geq0}\mathbb{E}_{\tilde{\bm{C}}}[\LInner(\tilde{\bm{C}}, \bm{\lambda})].
\]
Then, it holds that
\begin{align}
\sum_{t=1}^{T} \hLInner_t(\hbc^*_t, \hat{\bm{\lambda}}^*)&\leq \sum_{t=1}^{T} \mathbb{E}_{\tbc_t}[\hLInner_t(\tbc^*_t, \hat{\bm{\lambda}}^*)]\leq \sum_{t=1}^{T} \mathbb{E}_{\tbc_t}[\LInner_t(\tbc^*_t, \hat{\bm{\lambda}}^*)]+\frac{\|\hat{\bm{\lambda}}^*\|_1\cdot W_T}{T\beta_{\min}}=\mathbb{E}_{\tilde{\bm{C}}}[\LInner(\tilde{\bm{C}}^*, \hat{\bm{\lambda}}^*)]+\frac{\|\hat{\bm{\lambda}}^*\|_1\cdot W_T}{T\beta_{\min}}  \nonumber\\
&\leq \mathbb{E}_{\tilde{\bm{C}}}[\LInner(\tilde{\bm{C}}^*, \hat{\bm{\lambda}}^*)]+\frac{\|\hat{\bm{\lambda}}^*\|_1\cdot W_T}{T\beta_{\min}}=\OPTInner+\frac{\|\hat{\bm{\lambda}}^*\|_1\cdot W_T}{T\beta_{\min}}, \label{eqn:122003}
\end{align}
where the first inequality follows from definition of $(\hat{\bm{\lambda}^*}, (\hbc_t)_{t=1}^T)$, the second inequality follows from \eqref{eqn:121910}, the first equality follows from \eqref{eqn:121908} and the third inequality follows from the saddle-point condition of $(\bm{\lambda}^*, \tilde{\bm{C}}^*)$.

Plugging \eqref{eqn:122003} into \eqref{eqn:121918}, we have
\begin{equation}\label{eqn:122004}
\mathbb{E}_{\bm{\theta}\sim\bm{P}}\left[ \sum_{t=1}^{T} \left(p(\bm{c}_t)+f_{\theta_t}(\bm{x}_t)\right) \right]\leq \OPTInner+\frac{2\mu W_T}{T\beta_{\min}}+\mathbb{E}_{\bm{\theta}\sim\bm{P}}\left[\text{Reg}(T, \bm{\theta})\right]-\frac{\mu}{\beta_{\min}}\cdot\mathbb{E}_{\bm{\theta}\sim\bm{P}}\left[ d_T(\mathcal{A}, \bm{\theta}) \right].
\end{equation}
with $\mu=\|\hat{\bm{\lambda}}^*\|_1$.
From the non-negativity of $d_T(\mathcal{A}, \bm{\theta})$, we have
\[
\mathbb{E}_{\bm{\theta}\sim\bm{P}}\left[ \sum_{t=1}^{T} \left(p(\bm{c}_t)+f_{\theta_t}(\bm{x}_t)\right) \right]\leq \OPT+\frac{2\mu W_T}{T\beta_{\min}}+\mathbb{E}_{\bm{\theta}\sim\bm{P}}\left[\text{Reg}(T, \bm{\theta})\right].
\]
Using \Cref{thm:Hedgeregret} to bound $\mathbb{E}_{\bm{\theta}\sim\bm{P}}\left[\text{Reg}(T, \bm{\theta})\right]$, we have
\[
\mathbb{E}_{\bm{\theta}\sim\bm{P}}\left[ \sum_{t=1}^{T} \left(p(\bm{c}_t)+f_{\theta_t}(\bm{x}_t)\right) \right]\leq \OPT+\frac{2\mu W_T}{T\beta_{\min}}+\tilde{O}(\sqrt{T\cdot\log m})
\]
which completes our proof of \eqref{eqn:122001} by noting that $\mu=\alpha\cdot T$ for some constant $\alpha>0$.

We note that $(\bm{c}_t, \bm{x}_t)_{t=1}^T$ defines a feasible solution to
\begin{align}
 \OPT^{\delta}= \min & \sum_{t=1}^{T}\mathbb{E}_{\bm{c}_t, \bm{x}_t, \theta_t\sim P_t}\left[p(\bm{c}_t)+f_{\theta_t}(\bm{x}_t)\right] \label{lp:NOuterA}\\
  \mbox{s.t.} & \frac{1}{T}\cdot\sum_{t=1}^{T}\mathbb{E}_{\bc_t, \bm{x}_t, \theta_t\sim P_t}[\bm{g}_{\theta_t}(\bc_t, \bm{x}_t)]\leq\bm{\beta}+\delta\cdot\bm{e} \nonumber\\
  & \bm{x}_t\in\mathcal{K}(\theta_t, \bm{c}_t), \bm{c}_t\in\mathcal{C}, \forall t.\nonumber
\end{align}
with $\delta=\mathbb{E}_{\bm{\theta}\sim\bm{P}}[d_T(\mathcal{A}, \bm{\theta})]$ and $\bm{e}$ denotes a vector of all ones. We define another optimization problem by changing $P_t$ into $\hat{P}_t$ for each $t$,
\begin{align}
 \hOPT^{\hat{\delta}}= \min & \sum_{t=1}^{T}\mathbb{E}_{\bm{c}_t, \bm{x}_t, \theta_t\sim\hat{P}_t}\left[p(\bm{c}_t)+f_{\theta_t}(\bm{x}_t)\right] \label{lp:hNOuterA}\\
  \mbox{s.t.} & \frac{1}{T}\cdot\sum_{t=1}^{T}\mathbb{E}_{\bc_t, \bm{x}_t, \theta_t\sim\hat{P}_t}[\bm{g}_{\theta_t}(\bc_t, \bm{x}_t)]\leq\bm{\beta}+\hat{\delta}\cdot\bm{e} \nonumber\\
  & \bm{x}_t\in\mathcal{K}(\theta_t, \bm{c}_t), \bm{c}_t\in\mathcal{C}, \forall t.\nonumber
\end{align}
We have the following result that bounds the gap between $\OPT^{\delta}$ and $\hOPT^{\hat{\delta}}$, for some specific $\delta$ and $\hat{\delta}$.
\begin{claim}\label{claim:Bound}
For any $\delta\geq0$, if we set $\hat{\delta}=\delta+\frac{W_T}{T}$, then we have
\[
\hOPT^{\hat{\delta}}\leq\OPT^{\delta}+W_T.
\]
\end{claim}

If we regard $\hOPT^{\hat{\delta}}$ as a function over $\hat{\delta}$, then $\hOPT^{\hat{\delta}}$ is clearly a convex function over $\hat{\delta}$, where the proof follows the same spirit as the proof of \Cref{lem:ConvexL}. Moreover, note that
\[
\frac{d\hOPT^{\hat{\delta}=0}}{d\hat{\delta}}=\|\hat{\bm{\lambda}}^*\|_1\leq\mu.
\]
We have
\[
\hOPT=\hOPT^{0}\leq \hOPT^{\hat{\delta}}+  \mu\cdot \hat{\delta}
\]
for any $\hat{\delta}\geq0$. On the other hand, we have
\begin{equation}\label{eqn:010401}
\hOPT=\max_{\bm{\lambda}\geq0}\min_{\bm{c}_t\in\mathcal{C}} \hLInner(\bm{C}, \bm{\lambda})=\sum_{t=1}^{T} \hLInner_t(\hbc^*_t, \hat{\bm{\lambda}}^*)
\end{equation}
where the last equality follows from \Cref{lem:decompose}. Therefore, we now assume that $\mathbb{E}_{\bm{\theta}\sim\bm{P}}[d_T(\mathcal{A}, \bm{\theta})]\geq\frac{W_T}{T}$, otherwise \eqref{eqn:122002} directly holds. We have
\begin{equation}\label{eqn:121914}
\begin{aligned}
\sum_{t=1}^{T} \hLInner_t(\hbc^*_t, \hat{\bm{\lambda}}^*)&=\hOPT \leq \hOPT^{\mathbb{E}_{\bm{\theta}\sim\bm{P}}[d_T(\mathcal{A},\bm{\theta})]+\frac{W_T}{T}}+ \mu\cdot \left(\mathbb{E}_{\bm{\theta}\sim\bm{P}}[d_T(\mathcal{A},\bm{\theta})]+\frac{W_T}{T}\right)\\
&\leq \OPT^{\mathbb{E}_{\bm{\theta}\sim\bm{P}}[d_T(\mathcal{A},\bm{\theta})]}+ \mu\cdot \mathbb{E}_{\bm{\theta}\sim\bm{P}}[d_T(\mathcal{A},\bm{\theta})]+W_T\cdot(1+\frac{\mu}{T})\\
&\leq \sum_{t=1}^{T} \mathbb{E}_{\bm{\theta}\sim\bm{P}}\left[p(\bm{c}_t)+f_{\theta_t}(\bm{x}_t)\right]+\mu\cdot \mathbb{E}_{\bm{\theta}\sim\bm{P}}[d_T(\mathcal{A},\bm{\theta})]+W_T\cdot(1+\frac{\mu}{T})
\end{aligned}
\end{equation}
where the second inequality follows from \Cref{claim:Bound} by setting $\delta=\mathbb{E}_{\bm{\theta}\sim\bm{P}}[d_T(\mathcal{A},\bm{\theta})]$.
Plugging \eqref{eqn:121914} into \eqref{eqn:121918}, we have
\[\begin{aligned}
\sum_{t=1}^{T} \hLInner_t(\hbc^*_t, \hat{\bm{\lambda}}^*)\leq & \sum_{t=1}^{T} \hLInner_t(\hbc^*_t, \hat{\bm{\lambda}}^*)+\frac{\mu\cdot W_T}{T\beta_{\min}}+\mathbb{E}_{\bm{\theta}\sim\bm{P}}\left[\text{Reg}(T, \bm{\theta})\right]-\frac{\mu}{\beta_{\min}}\cdot\mathbb{E}_{\bm{\theta}\sim\bm{P}}\left[ d_T(\mathcal{A}, \bm{\theta}) \right]\\
&+\mu\cdot \mathbb{E}_{\bm{\theta}\sim\bm{P}}[d_T(\mathcal{A},\bm{\theta})]+W_T\cdot(1+\frac{\mu}{T}),
\end{aligned}\]
which implies
\[
\mu\cdot\left(\frac{1}{\beta_{\min}}-1\right)\cdot \mathbb{E}_{\bm{\theta}\sim\bm{P}}[ d_T(\mathcal{A},\bm{\theta})]\leq \mathbb{E}_{\bm{\theta}\sim\bm{P}}\left[\text{Reg}(T, \bm{\theta})\right]+\frac{2\mu W_T}{T\beta_{\min}}++W_T\cdot(1+\frac{\mu}{T}).
\]
which completes our final proof by noting that $\mu=\alpha\cdot T$ for some constant $\alpha>0$.
\end{myproof}

\begin{myproof}[Proof of \Cref{claim:Bound}]
Denote by $(\tilde{\bm{c}}_t, \tilde{\bm{x}}_t)_{t=1}^T$ one optimal solution to $\OPT^{\delta}$. Then, from the definition of $W_T$, we have that
\[
\frac{1}{T}\cdot\sum_{t=1}^{T}\mathbb{E}_{\tilde{\bc}_t, \tilde{\bm{x}}_t, \theta_t\sim\hat{P}_t}[\bm{g}_{\theta_t}(\bc_t, \bm{x}_t)]\leq \frac{1}{T}\cdot\sum_{t=1}^{T}\mathbb{E}_{\tilde{\bc}_t, \tilde{\bm{x}}_t, \theta_t\sim P_t}[\bm{g}_{\theta_t}(\bc_t, \bm{x}_t)]+\frac{W_T}{T}.
\]
Therefore, we know that $(\tilde{\bm{c}}_t, \tilde{\bm{x}}_t)_{t=1}^T$ is a feasible solution to $\hOPT^{\hat{\delta}}$. On the other hand, from the definition of $W_T$, we know that
\[
\hOPT^{\hat{\delta}}\leq\sum_{t=1}^{T}\mathbb{E}_{\tilde{\bm{c}}_t, \tilde{\bm{x}}_t, \theta_t\sim\hat{P}_t}\left[p(\tilde{\bm{c}}_t)+f_{\theta_t}(\tilde{\bm{x}}_t)\right]\leq \sum_{t=1}^{T}\mathbb{E}_{\tilde{\bm{c}}_t, \tilde{\bm{x}}_t, \theta_t\sim\hat{P}_t}\left[p(\tilde{\bm{c}}_t)+f_{\theta_t}(\tilde{\bm{x}}_t)\right]+W_T=\OPT^{\delta}+W_T
\]
by noting that the distribution of $\tilde{\bm{c}}_t$ must be independent of $\theta_t$ for each $t$, which completes our proof.
\end{myproof}

Notably, the inaccuracy of the prior estimates would result in a constraint violation that scales as $O(\frac{1}{\sqrt{T}}+\frac{W_T}{T})$ as shown in \Cref{thm:NStaRegretISP}. Therefore, as long as $W_T$ scales sublinearly in $T$, which guarantees a sublinear regret bound following \eqref{eqn:122001}, the constraint violation of our \Cref{alg:IAL} also scales as $o(1)$, implying that the solutions generated by \Cref{alg:IAL} is asymptotically feasible. Generating asymptotically feasible solutions is standard in OCOwC literature \citep{jenatton2016adaptive, neely2017online, yuan2018online,yi2021regret}. Moreover, though in the derivation, we use Wasserstein distance in the definition \eqref{eqn:Wasserdist}, we note that \Cref{thm:NStaRegretISP} in fact holds for general definition of $W_T$ where $W(P_t, \hat{P}_t)$ in \eqref{eqn:Wasserdist} can be defined by any other measure. For the packing problem where the first-stage actions is de-activated, it has been shown in \cite{balseiro2023robust} that instead of knowing the entire prior estimate $\hat{\bm{P}}$, having a single sample of $\hat{P}_t$ for each $t\in[T]$ is enough to obtain the $O(\sqrt{T}+W_T)$ regret. It is interesting to see how many samples are needed to obtain the same regret for the general two-stage problem and we leave this question for future research.

\section{Numerical Experiments}\label{sec:numerical}

In this section, we conduct numerical experiments to test the performance of our algorithms empirically.
We try two sets of experiments for the resource allocation problems.
One deals with the resource capacity constraints (packing constraints from a mathematical optimization view), where the long-term constraints take the formulation of $\sum_{t=1}^{T}\bm{g}(\bx_t, \bm{\theta}_t)\leq T\cdot\bm{\beta}$. The other set of experiments deals with the service level constraints (covering constraints from a mathematical optimization view), where the long-term constraints take the formulation of $\sum_{t=1}^{T}\bm{g}(\bx_t, \bm{\theta}_t)\geq T\cdot\bm{\beta}$. Note that the service level constraints (or the fairness constraints \citep{kearns2018preventing}) are widely studied in the literature (e.g. \cite{hou2009theory}), and we develop our algorithms to handle this type of covering constraints. The algorithmic development for the second set is described in \Cref{sec:Covering} in the supplementary material. In both of our experiments, we set the functions $\bm{g}(\bx_t, \bm{\theta}_t)$ to be linear functions over $\bx_t$.

\noindent\textbf{Experiment 1.} The first set deals with packing constraints. The offline problem can be formulated as follows:
\begin{align*}
&\max~~\sum_{t=1}^{T} c_t\\
&~~\mbox{s.t.}~~~\sum_{t=1}^{T}x_{i,t}\leq T\cdot \beta_i, ~~\forall i=1,\dots, 4\\
&~~~~~~~~~~x_{i,t}\leq D_{i,t,},~~\forall i, \forall t\\
&~~~~~~~~~~\sum_{i=1}^{4}x_{i,t}\geq \min\{ c_t, \sum_{i=1}^{4}D_{i,t} \},~~\forall t\\
&~~~~~~~~~~x_{i,t}\geq0, c_t\leq C,~~\forall i, \forall t.
\end{align*}
We do the numerical experiment under the following settings. Consider a resource allocation problem where there is 4 resources to be allocated. At each period, the decision maker needs to first decide a budget $c_t$ which restricts the total amount of units to be invested to each resources, and then, after the demand is realized, the decision maker needs to allocated the budget to satisfy the demand for each resource. The demand for each resource is normally distributed at each period (truncated at 0). We set the mean parameter $\mu_0=5$ and the standard deviation parameter $\sigma_0=10/3$. We set the vector $\bm{\beta}=[0.95, 0.90, 0.85, 0.80]$. We assume DAL is blind to the distributions while IAL knows the distributions. We test the performance of the IAL algorithm and DAL algorithm in the following four cases:
\begin{itemize}
  \item a. Stationary distribution case: demands of the four resources follow the same distribution $\mathcal{N}(k_0\mu_0, \sigma_0)$ with $k_0=2$.
  \item b. Non-stationary distribution case 1: We divide the whole horizon into two intervals: $[1, T/2]$ and $[T/2+1, T]$. During each interval, we sample the demands independently following the same distribution $\mathcal{N}(k_1\mu_0, \sigma_0)$ with $k_1=1, 3$ for the two intervals.
  \item c. Non-stationary distribution case 1: We divide the whole horizon into two intervals: $[1, T/2]$ and $[T/2+1, T]$. During each interval, we sample the demands independently following the same distribution $\mathcal{N}(k_2\mu_0, \sigma_0)$ with $k_2=3, 1$ for the two intervals.
  \item d. Non-stationary distribution case 3: We divide the whole horizon into five intervals: $[1, T/5]$, $[T/5+1, 2T/5]$, $[2T/5+1, 3T/5]$, $[3T/5+1, 4T/5]$ and $[4T/5+1, T]$. During each interval, we sample the demands independently following the same distribution $\mathcal{N}(k_3\mu_0, \sigma_0)$ with $k_3=1, 2, 3, 2, 1$ for the five intervals.
\end{itemize}
\textbf{Results and interpretations.} Note that the extent of non-stationarity is the same for the non-stationary case 1 and case 2. The difference is that the distribution of the two intervals is exchanged in case 1 and case 2. The extent of non-stationarity is largest for the non-stationary case 3. The numerical results are presented in \Cref{tab:01}. On one hand, for the stationary case, the performances of both DAL and IAL are good, with a relative regret within $4\%$. On the other hand, as we can see, the performance of DAL decays from situation (a) up to situation (d), as the non-stationarity of the underlying distributions gets larger and larger. However, the performance of IAL remains relatively stable because the non-stationarity of the underlying distributions has been well incorporated in the algorithmic design, which illustrates the effectiveness of IAL in a non-stationary environment.
\begin{table}[!h]
  \centering
  \begin{tabular}{|c|c|c|c|c|}
    \hline
     & stationary case & non-stationary case 1 & non-stationary case 2 & non-stationary case 3 \\
     \hline
    DAL & $3.43\%$ & $8.26\%$ & $8.26\%$ & $11.02\%$ \\
    \hline
    IAL & $3.45\%$  & $5.84\%$ & $5.86\%$ & $6.54\%$ \\
    \hline
  \end{tabular}
  \caption{The relative regret of DAL and IAL algorithms with capacity (packing) constraints}
  \label{tab:01}
\end{table}

\begin{figure}[!h]
\centering
\begin{subfigure}{0.45\textwidth}
    \includegraphics[width=\textwidth]{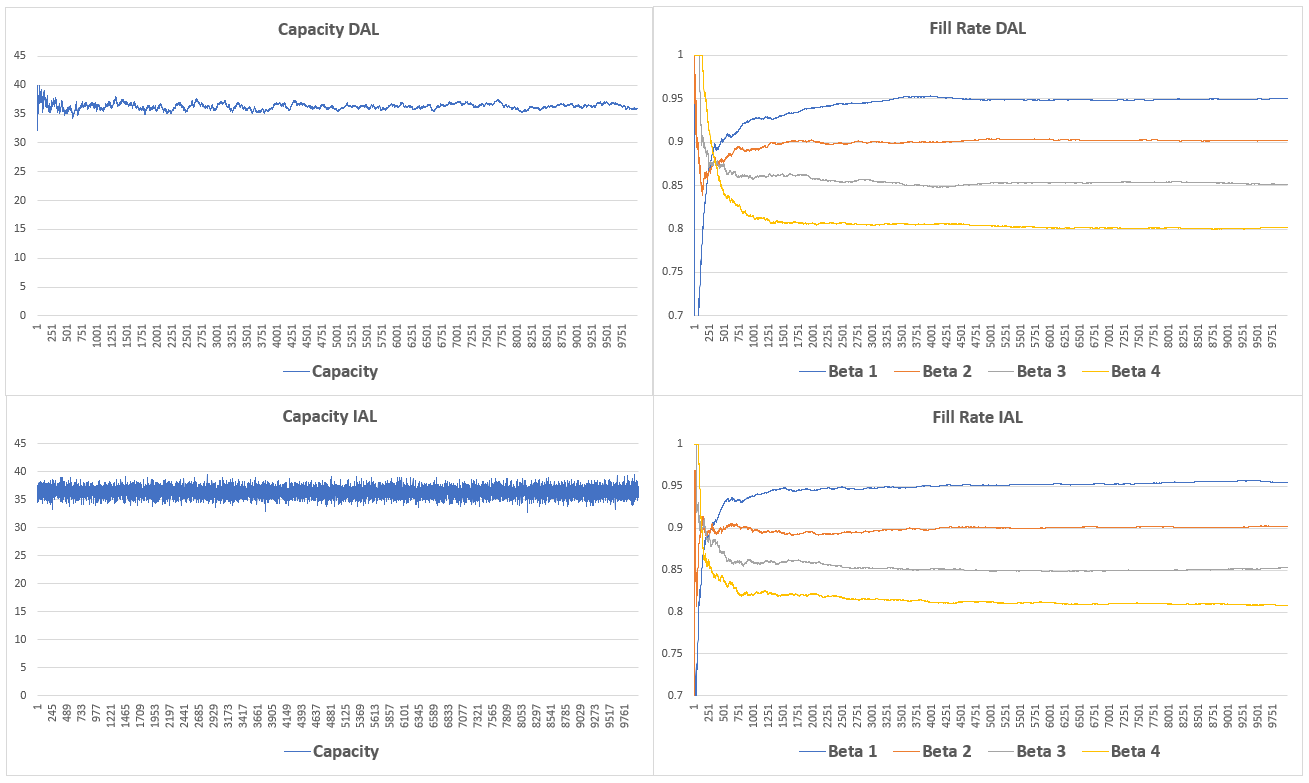}
    \caption{Results for the stationary case.}
    \label{fig:a}
\end{subfigure}
\hfill
\begin{subfigure}{0.45\textwidth}
    \includegraphics[width=\textwidth]{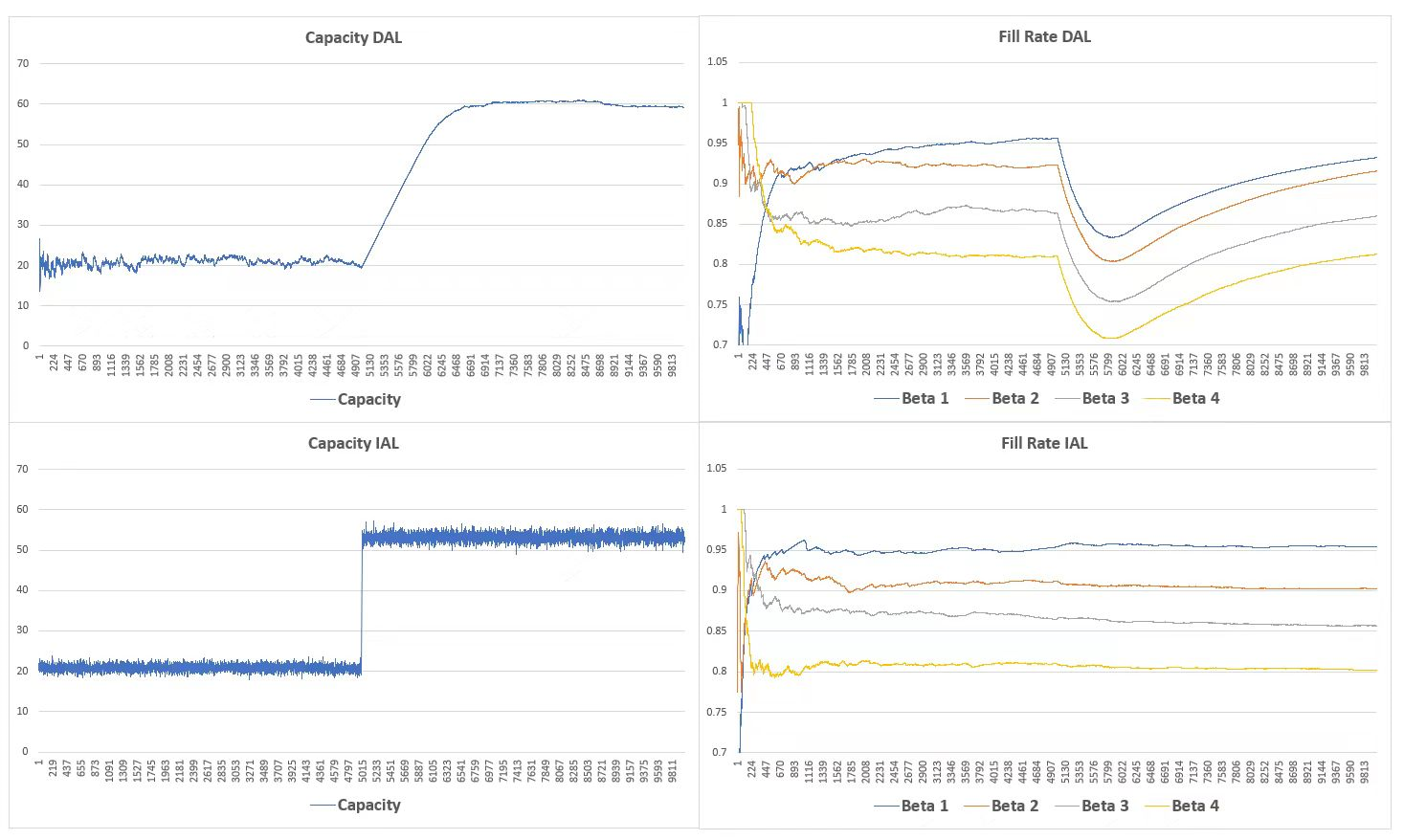}
    \caption{Results for the non-stationary case 1.}
    \label{fig:b}
\end{subfigure}
\hfill
\begin{subfigure}{0.45\textwidth}
    \includegraphics[width=\textwidth]{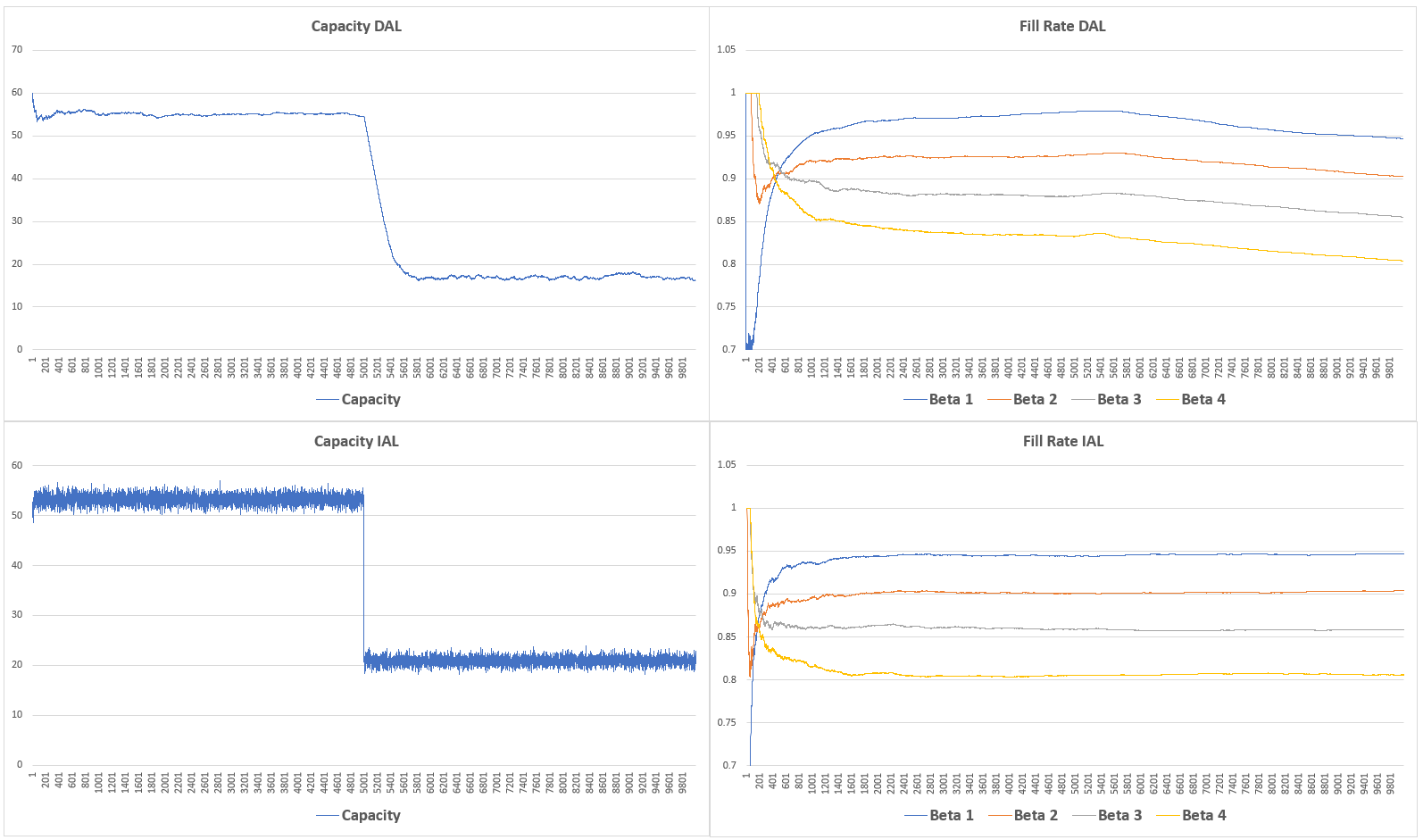}
    \caption{Results for the non-stationary case 2.}
    \label{fig:c}
\end{subfigure}
\hfill
\begin{subfigure}{0.45\textwidth}
    \includegraphics[width=\textwidth]{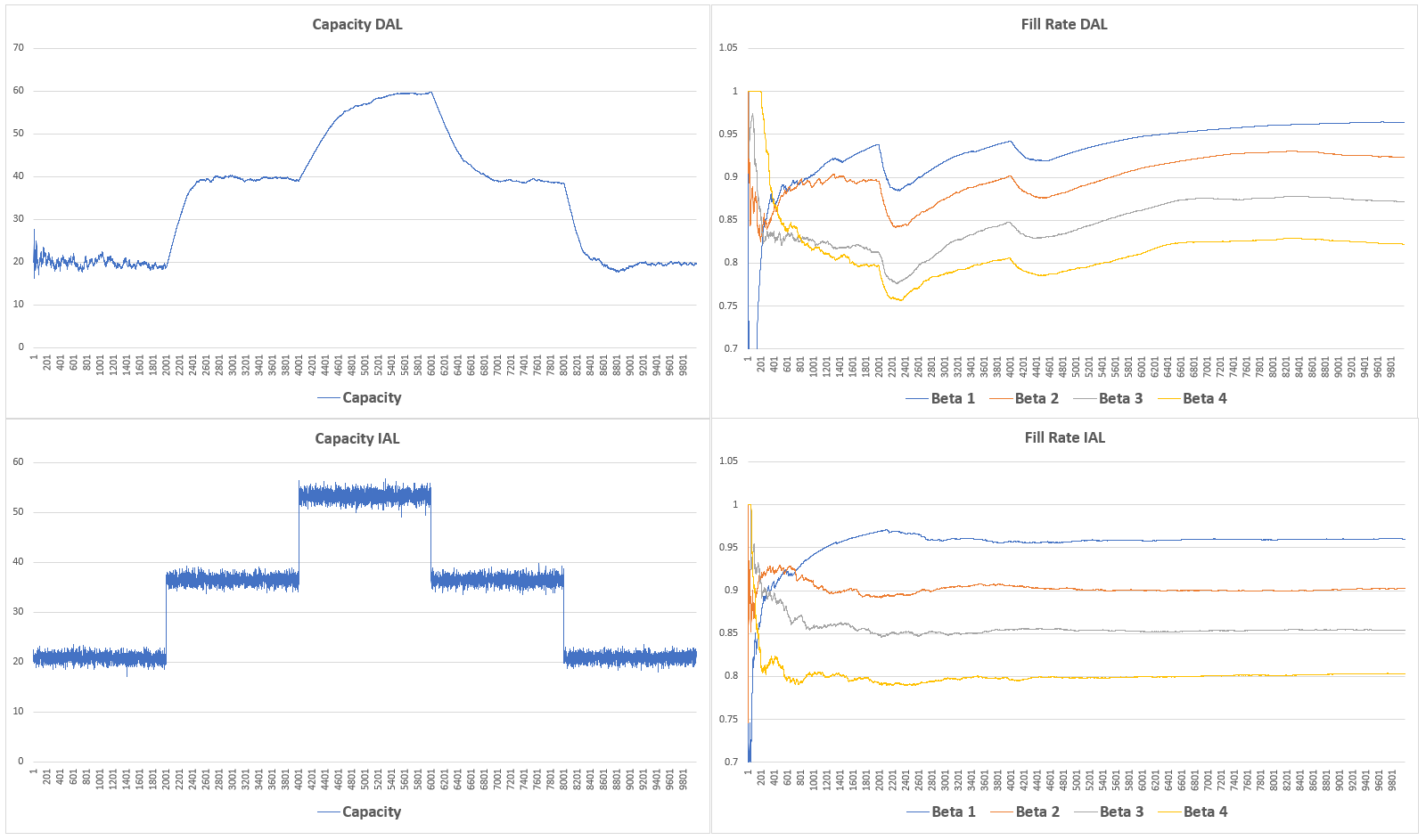}
    \caption{Results for the non-stationary case 3.}
    \label{fig:d}
\end{subfigure}

\caption{Numerical results of DAL and IAL algorithms with service level (covering) constraints.}
\label{fig:figures}
\end{figure}

\noindent\textbf{Experiment 2.}  The second set considers the service level constraints, namely, the covering constraints. We consider the four cases which are exactly the same as the cases described in experiment 1. The other parameters are also set as the same as those in experiment 1, except that the long-term constraints now become $\sum_{t=1}^{T}\bm{g}(\bx_t, \bm{\theta}_t)\geq T\cdot\bm{\beta}$.

\noindent\textbf{Results and interpretations.} The numerical results are given in \Cref{fig:figures}. In each of the subfigures, corresponding to each case, we plot 4 graphs capturing how the budget $c_t$
 is determined and how the service level constraints are satisfied for each DAL and IAL during the horizon of $T=10000$
 period. As we can see, both IAL and DAL converge rather quickly under the stationary case, in that the budget $c_t$
 remains stable and the target service levels are achieved after about $1000$
 periods, as plotted in \Cref{fig:a}. However, as we add non-stationarity in \Cref{fig:b}, \Cref{fig:c}, and \Cref{fig:d}, the performance of DAL decays in that the budget $c_t$
 cannot capture the non-stationarity, and the achieved service levels are not stable. In contrast, for IAL, the budget $c_t$
 changes quite quickly as long as the underlying distribution shifts. Moreover, even though the distributions changed during the horizon, the achieved service levels of IAL remain stable and the targets are reached. All these numerical results correspond to our theoretical finding and illustrate the benefits of IAL under non-stationarity.

\section{Summary}\label{sec:summary}
This paper proposes and studies the problem of bounding regret for online two-stage stochastic optimization with long-term constraints. The main contribution is an algorithmic framework that develops new algorithms via adversarial learning algorithms. The framework is applied to various setting. For the stationary setting, the resulted DAL algorithm is shown to achieve a sublinear regret, with a performance robust to adversarial corruptions. For the non-stationary (adversarial) setting, a modified IAL algorithm is developed, with the help of prior estimates. The sublinear regret can also be acheived by IAL algorithm as long as the cumulative inaccuracy of the prior estimates is sublinear. We explore extensions of our approach and show that our results still hold if the convexity requirements in \Cref{assump:main} are removed and if we have covering constraints in our model.

\bibliographystyle{abbrvnat}
\bibliography{bibliography}

\clearpage

%
%
%

\begin{APPENDICES}
\crefalias{section}{appendix}

\section{Extensions}\label{sec:extensions}

\subsection{Non-convex Objective and Non-concave Constraints}

Note that in \Cref{assump:main}, we require a convexity condition where the function $p(\cdot)$ is a convex function with $p(\bm{0})=0$ and $\mathcal{C}$ is a compact and convex set which contains $\bm{0}$. Also, the functions $f_{\theta}(\cdot)$ need to be convex in $\bm{x}$ and $\bm{g}_{\theta}(\cdot, \cdot)$ need to be convex in both $\bc$ and $\bm{x}$, for any $\theta$. In this section, we explore whether these convexity conditions can be removed with further characterization of our model.

We now show that when $\mathcal{C}$ contains only a finite number of elements, we can remove all the convexity requirements in \Cref{assump:main} and we still obtain a $\tilde{O}(\sqrt{T})$ regret bound for the setting in \Cref{sec:stationary}, a $\tilde{O}(\sqrt{T}+W_T)$ regret bound for the setting in \Cref{sec:adversarial}. As for the setting in \Cref{sec:nonstationary}, we can obtain the same $\tilde{O}(\sqrt{T}+W_T)$ regret bound for \textit{arbitrary} $\mathcal{C}$, as long as the minimization step $\bm{c}_t=\text{argmin}_{\bm{c}\in\mathcal{C}}\hLInner_t(\bm{c}, \mu\cdot\bm{e}_{i_t})$ can be efficiently solved in \Cref{alg:IAL}.

We now assume $\mathcal{C}$ has a finite support and remove all the convexity requirements in \Cref{assump:main}. In order to recover the results in \Cref{sec:stationary} and \Cref{sec:adversarial}, we regard each element in $\mathcal{C}$ as an expert and we apply expert algorithms to decide $\bc_t$ for each period $t$. To be more concrete, note that for the first-stage decision, we only have \textit{bandit-feedback}, i.e., we can only observe the stochastic outcome of selecting $\bc_t$ instead of other elements in the set $\mathcal{C}$. Therefore, we can apply EXP3 algorithm \citep{auer2002nonstochastic} as $\ALG_1$. Still, following the same procedure as the proof of \Cref{thm:RegretISP} and the more general \Cref{thm:CorruptRegretISP}, we can reduce the regret bound of \Cref{alg:DAL} into the regret bound of $\ALG_1$, $\ALG_2$, and an additional $O(W_T)$ term if adversarial corruption exists. The above argument is formalized in the following theorem.
\begin{theorem}\label{thm:DiscreteC}
Denote by $\pi$ \Cref{alg:DAL} with input $\mu=T$. Then, if $\mathcal{C}$ constains only a finite number of elements, denoted by $K$, with condition c in \Cref{assump:main}, it holds that
\begin{align}
\text{Regret}^{\text{c}}(\pi, T)=\mathbb{E}_{\bm{\theta}\sim\bm{P}}[\ALG(\pi,\bm{\theta}^c)]-\mathbb{E}_{\bm{\theta}\sim\bm{P}}[\ALG(\pi^*,\bm{\theta}^c)] \leq &\tilde{O}(\sqrt{K\cdot T})+\tilde{O}(\sqrt{T\cdot\log m})\nonumber\\
&+O( \mathbb{E}_{\bm{\theta}\sim\bm{P}}[W(\bm{\theta})] ) , \label{eqn:012501}
\end{align}
if EXP3 is selected as $\ALG_1$ and Hedge is selected as $\ALG_2$. Here, $\mathbb{E}_{\bm{\theta}\sim\bm{P}}[W(\bm{\theta})]$ denotes the expectation of the total number of corruptions, with $W(\bm{\theta})$ defined in \eqref{eqn:Advercorrup}.
\end{theorem}

We now consider the setting in \Cref{sec:nonstationary} and we show that the $\tilde{O}(\sqrt{T}+W_T)$ regret bound holds for \Cref{alg:IAL} without the convexity requirements in \Cref{assump:main}, for arbitrary $\mathcal{C}$ as long as the minimization step $\bm{c}_t=\text{argmin}_{\bm{c}\in\mathcal{C}}\hLInner_t(\bm{c}, \mu\cdot\bm{e}_{i_t})$ can be efficiently solved. The key idea is to regard the first-stage decision at each period as a distribution over $\mathcal{C}$, denoted by $\tbc_t$. Then, $\mathbb{E}_{\tilde{\bm{C}}}\left[\hat{L}(\tilde{\bm{C}}, \bm{\lambda})\right]$ would be a convex function over the distribution $\tilde{\bm{C}}$. Moreover, for each fixed $\bm{\lambda}$, note that selecting the worst distribution $\tilde{\bm{C}}$ to minimize $\mathbb{E}_{\tilde{\bm{C}}}\left[\hat{L}(\tilde{\bm{C}}, \bm{\lambda})\right]$ can be reduced equivalent to selecting the worst deterministic $\bm{C}$ to minimize $\hat{L}(\bm{C}, \bm{\lambda})$. Therefore, every step of the proof of \Cref{lem:decompose} and \Cref{thm:NStaRegretISP} would follow by replacing $\bc_t$ into the distribution $\tbc_t$ for each $t\in[T]$. We have the following result.
\begin{theorem}\label{thm:NconvexNStaRegretISP}
Denote by $\pi$ \Cref{alg:IAL} with input $\mu=\|\hat{\bm{\lambda}}^*\|_{\infty}=\alpha\cdot T$ for some constant $\alpha>0$. Denote by $W_T=\sum_{t=1}^{T}W(\hat{P}_t, P_t)$ the total measure of inaccuracy, with $W(\hat{P}_t, P_t)$ defined in \eqref{eqn:Wasserdist}. Then, under condition c of \Cref{assump:main}, for arbitrary $\mathcal{C}$, the regret enjoys the upper bound
\begin{align}
\text{Regret}(\pi, T)=&\mathbb{E}_{\bm{\theta}\sim\bm{P}}[\ALG(\pi,\bm{\theta})]-\mathbb{E}_{\bm{\theta}\sim\bm{P}}[\ALG(\pi^*,\bm{\theta})]
\leq \tilde{O}(\sqrt{T\cdot\log m})+O( W_T )  \label{eqn:012509}
\end{align}
if Hedge is selected as $\ALGD$. Moreover, we have
\begin{align}
\frac{1}{T}\sum_{t=1}^{T}g_{i,\theta_t}(\bc_t, \bm{x}_t)-\beta_i\leq \tilde{O}\left(\sqrt{\frac{\log m}{T}}\right)+O\left(\frac{W_T}{ T}\right), \label{eqn:012510}
\end{align}
for each $i\in[m]$.
\end{theorem}

\subsection{Incorporating Covering Constraints}\label{sec:Covering}

In this section, we discuss how to incorporate covering constraints into our model. Note that in the previous section, we let the long-term constraints $\frac{1}{T}\cdot\sum_{t=1}^{T}\bx_t\in\mathcal{B}(\bm{C}, \bm{\theta})$ be characterized as follows,
\begin{equation}\label{eqn:012511}
  \frac{1}{T}\cdot\sum_{t=1}^{T}\bm{g}_{\theta_t}(\bc_t, \bm{x}_t)\leq\bm{\beta},
\end{equation}
with $\bm{\beta}\in(0,1)^m$, which corresponds to packing constraints since both $\bm{g}(\cdot)$ and $\bm{\beta}$ are non-negative. For the case with covering constraints, the long-term constraints $\frac{1}{T}\cdot\sum_{t=1}^{T}\bx_t\in\mathcal{B}(\bm{C}, \bm{\theta})$ would enjoy the following characterization
\begin{equation}\label{eqn:012512}
  \frac{1}{T}\cdot\sum_{t=1}^{T}\bm{g}_{\theta_t}(\bc_t, \bm{x}_t)\geq\bm{\beta},
\end{equation}
with $\bm{\beta}\in(0,1)^m$. We now show that all our previous results hold for the packing constraints \eqref{eqn:012512}, by simply changing the input parameter $\mu$ in \Cref{alg:DAL} and \Cref{alg:IAL}. We illustrate through the stationary setting consiered in \Cref{sec:stationary} where $P_t=P$ for each $t\in[T]$.

Now, \eqref{lp:OPT} enjoys the following new formulation.
\begin{align}
 \OPTOuter= \min & \sum_{t=1}^{T}\mathbb{E}_{\tbc_t, \tbx_t, \theta_t}\left[p(\tbc_t)+f_{\theta_t}(\tbx_t)\right] \label{lp:Outer}\\
  \mbox{s.t.} & \frac{1}{T}\cdot\sum_{t=1}^{T}\mathbb{E}_{\tbc_t, \tbx_t, \theta_t}[\bm{g}_{\theta_t}(\tbc_t, \tbx_t)]\geq\bm{\beta} \nonumber\\
  & \tbx_t\in\mathcal{K}(\theta_t, \tbc_t), \tbc_t\in\mathcal{C}, \forall t.\nonumber
\end{align}
The Lagrangian dual of $\OPTOuter$ \eqref{lp:Outer} can be formulated as follows.
\begin{equation}\label{lp:Dualouter}
  \max_{\bm{\lambda}\geq0}\min_{\tbc_t\in\mathcal{C}} \LOuter(\bm{C}, \bm{\lambda})=\sum_{t=1}^{T}\mathbb{E}\left[p(\tbc_t)+\frac{1}{T}\cdot\sum_{i=1}^{m}\lambda_i+\min_{\tbx_t\in\mathcal{K}(\theta_t,\tbc_t)}f_{\theta_t}(\tbx_t)-\sum_{i=1}^{m}\frac{\lambda_i\cdot g_{i,\theta_t}(\tbc_t, \tbx_t)}{T\cdot\beta_i}\right].
\end{equation}
Similar to \eqref{eqn:LISPstationary}, we have the following formulation for $\bLInner(\bm{c}, \bm{\lambda}, \theta)$ as the single-period decomposition of $ \LOuter(\bm{C}, \bm{\lambda})$.
\begin{equation}\label{eqn:LOSPstationary}
\bLOuter(\bm{c}, \bm{\lambda}, \theta)=p(\bm{c})+\frac{1}{T}\cdot\sum_{i=1}^{m}\lambda_i+\min_{\bm{x}\in\mathcal{K}(\theta,\bm{c})}\left\{f_{\theta}(\bm{x})-\sum_{i=1}^{m}\frac{\lambda_i\cdot g_{i,\theta}(\bc, \bm{x})}{T\cdot\beta_i}\right\}.
\end{equation}
Following the same procedure as the proof of \Cref{lem:StaDual}, we obtain the following result.
\begin{lemma}\label{lem:NEWStaDual}
Under the stationary setting where $P_t=P$ for each $t\in[T]$, it holds that
\begin{equation*}
\max_{\bm{\lambda}\geq0}\min_{\tbc_t\in\mathcal{C}} \mathbb{E}_{\tilde{\bm{C}}}[\LOuter(\bm{C}, \bm{\lambda})]=\max_{\bm{\lambda}\geq0}\min_{\tbc\in\mathcal{C}} T\cdot\mathbb{E}_{\tbc, \theta\sim P}\left[\bLOuter(\tbc, \bm{\lambda}, \theta)\right].
\end{equation*}
\end{lemma}
Therefore, we can still regard solving the dual problem \eqref{lp:Dualouter} as a procedure of solving the repeated zero-sum games, where player 1 chooses the first-stage decision $\bc_t$ and player 2 chooses one long term constraint $i_t\in[m]$. We still apply \Cref{alg:DAL} with the new definition
\begin{equation}\label{eqn:defOutOSP}
\bLOuter_{i}(\bm{c}_t, \mu\cdot\bm{e}_{i_t},\theta_t)=p(\bm{c}_t)+\frac{\mu}{T}+f_{\theta_t}(\bm{x}_t)-\frac{\mu\cdot g_{i,\theta_t}(\bc_t, \bm{x}_t)}{T\cdot\beta_i}.
\end{equation}
However, in order to incorporate covering constraints, we need to adjust the value of the scaling factor $\mu$ and we only terminate our algorithm after the entire horizon has been run out. In fact, we need an upper bound (arbitrary upper bound suffices) over the $l_1$ norm of the optimal dual variable $\bm{\lambda}^*$ of \eqref{lp:Dualouter} to serve as the scaling factor $\mu$ and in practice, we can spend the first $\sqrt{T}$ time periods to construct an upper confidence interval of $\bm{\lambda}^*$ to serve as $\mu$ without influencing the order of the regret bound.

\begin{theorem}\label{thm:RegretOSP}
Denote by $\pi$ \Cref{alg:DAL} with input $\mu$ being an upper bound of $\|\bm{\lambda}^*\|_1$ where $\bm{\lambda}^*$ is the optimal dual variable of \eqref{lp:Dualouter}. Then, under \Cref{assump:main}, $\mu=\alpha\cdot T$ for some constant $\alpha>0$ and if OGD is selected as $\ALG_1$ and Hedge is selected as $\ALG_2$, the regret enjoys the upper bound
\begin{equation}\label{eqn:121501}
\text{Regret}^{\text{TSC}}(\pi, T)\leq \tilde{O}((G+F)\cdot\sqrt{T})+\tilde{O}(\sqrt{T\cdot\log m})
\end{equation}
and moreover, we have
\begin{equation}\label{eqn:121502}
\beta_i-\frac{1}{T}\cdot\sum_{t=1}^{T}g_{i,\theta_t}(\bc_t, \bm{x}_t)\leq \tilde{O}\left(\frac{G+F}{\alpha\sqrt{T}}\right)+\tilde{O}\left(\frac{\sqrt{\log m}}{\alpha\cdot\sqrt{T}}\right) ,~~\forall i\in[m].
\end{equation}
\end{theorem}

\section{Regret Bounds for Adversarial Learning}

In this section, we present the implementation details of two adversarial learning algorithms, OGD and Hedge, that will be used as algorithmic subroutines in our \Cref{alg:DAL} and \Cref{alg:IAL}, as well as their regret analysis.

OGD is an algorithm to be executed in a finite horizon of $T$ periods, and at each period $t$, OGD selects an action $\bm{c}_t\in\mathcal{C}$, receives an adversarial chosen cost function $h_t(\cdot)$ afterwards, and incurs a cost $h_t(\bm{c}_t)$. OGD is designed to minimize the regret
\[
\text{Reg}_{\text{OGD}}(T)=\sum_{t=1}^{T}h_t(\bm{c}_t)-\min_{\bm{c}\in\mathcal{C}}\sum_{t=1}^{T}h_t(\bm{c}).
\]
The implementation of OGD is described in \Cref{alg:OGD}. In our problem, $h_t=\bar{L}^{\text{ISP}}_i(\bm{c}_t,\mu\cdot\bm{e}_{i_t},\theta_t)$ for ISP and $h_t=\bar{L}^{\text{OSP}}_i(\bm{c}_t,\mu\cdot\bm{e}_{i_t},\theta_t)$ for OSP. In order to obtain a subgradient, one can compute the optimal dual variable of the inner minimization problem of \eqref{eqn:LISPstationary}. For example, the inner minimization problem is
\begin{align}
  \min &~~~ f_{\theta_t}(\bm{x})+\frac{\mu\cdot g_{i_t,\theta_t}(\bc, \bm{x})}{T\cdot\beta_{i_t}} \label{lp:InnerISP}\\
  \mbox{s.t.} &~~~ B_{\theta_t}\bm{x}\leq \bm{c}_t\nonumber\\
  &~~~ \bm{x}\geq0 \nonumber
\end{align}
and the Lagrangian dual problem of \eqref{lp:InnerISP} is
\[
\max_{\bm{\gamma}\leq0} \bm{\gamma}^\top\bm{c_t}+\min_{\bm{x}\geq0}f_{\theta_t}(\bm{x})-\bm{\gamma}^\top B_{\theta_t}\bm{x}+\frac{\mu\cdot g_{i_t,\theta_t}(\bc_t, \bm{x})}{T\cdot\beta_{i_t}}.
\]
The optimal dual solution $\bm{\gamma}_t^*$ can be computed from the above minimax problem and we know that
\[
\nabla h_t(\bm{c}_t)=\bm{\gamma}_t^*+\frac{\mu\cdot \nabla_{\bc_t} g_{i_t,\theta_t}(\bc_t, \bm{x})}{T\cdot\beta_{i_t}}.
\]
Clearly when $\mu=a\cdot T$ for a constant $a>0$, we have $\|\bm{\gamma}_t^*\|_2\leq G$ for some constant $G$ that depends only on (the upper bound of gradients of) $\{f_{\theta}, \bm{g}_{\theta}\}_{\forall \theta}$, the minimum positive element of $B_{\theta}$ for all $\theta$, and $\max_{i\in[m]}\{\frac{1}{\beta_i}\}$. The regret bound of OGD is as follows.
\begin{theorem}[Theorem 1 of \cite{zinkevich2003online}]\label{thm:OGDregret}
If $\eta_t=\frac{1}{\sqrt{t}}$, then it holds that
\[
\text{Reg}_{\text{OGD}}(T)\leq O\left((G+F)\cdot\sqrt{T}\right)
\]
where $F$ is an upper bound of the diameter of the set $\mathcal{C}$ and $G$ is a constant that depends only on (the upper bound of gradients of) $\{f_{\theta}, \bm{g}_{\theta}\}_{\forall \theta}$, the minimum positive element of $B_{\theta}$ for all $\theta$, and $\max_{i\in[m]}\{\frac{1}{\beta_i}\}$.
\end{theorem}

\begin{algorithm}[tb]
\caption{Online Gradient Descent (OGD) algorithm}
\label{alg:OGD}
\begin{algorithmic}
\STATE {\bfseries Input:} the step size $\eta_t$ for each $t\in[T]$.
\STATE Initially set an arbitrarily $\bm{c}_1\in\mathcal{C}$.
\FOR{$t=1,\dots,T$}
\STATE $\bm{1}.$ Take the action $\bm{c}_t$.
\STATE $\bm{2}.$ Observe the cost function $h_t(\cdot)$.
\STATE $\bm{3}.$ Update action
\[
\bm{c}_{t+1}=\mathcal{P}_{\mathcal{C}}\left(\bm{c}_t-\eta_t\cdot \nabla h_t(\bm{c}_t)\right)
\]
where $\nabla h_t(\bm{c}_t)$ denotes a subgradient of $h_t$ at $\bm{c}_t$ and $\mathcal{P}_{\mathcal{C}}$ denotes a projection to the set $\mathcal{C}$.
\ENDFOR
\end{algorithmic}
\end{algorithm}

The Hedge algorithm is used to solve the expert problem in a finite horizon of $T$ periods. There are $m$ experts and at each period $t$, Hedge will select one expert $i_t\in[m]$ ($i_t$ can be randomly chosen), observe the reward vector $\bm{l}_t\in\mathbb{R}^m$ afterwards, and obtain an reward $l_{i_t,t}$. Hedge is designed to minimize the regret
\[
\text{Reg}_{\text{Hedge}}(T)=\max_{i\in[m]}\sum_{t=1}^{T}l_{i,t}-\sum_{t=1}^{T}\mathbb{E}_{i_t}[l_{i_t, t}].
\]
The Hedge algorithm is described in \Cref{alg:Hedge}. In our problem, for ISP, we have
\[
l_{i,t}=\bar{L}_{i}(\bm{c}_t, \mu\cdot\bm{e}_{i_t},\theta_t).
\]
Under \Cref{assump:main}, when $\mu=\alpha\cdot T$ for a constant $\alpha$, we know that there exists a constant $\delta>0$ such that $|l_{i,t}|\leq\delta$, for all $i\in[m]$ and $t\in[T]$. Here, $\delta$ depends on $\max_{i\in[m]}\{\frac{1}{\beta_i}\}$.
\begin{theorem}[from Theorem 2 in \cite{freund1997decision}]\label{thm:Hedgeregret}
If $\eps=\sqrt{\frac{\log m}{T}}$, then it holds that
\[
\text{Reg}_{\text{Hedge}}(T)\leq \tilde{O}(\sqrt{T\cdot\log (m)})
\]
where the constant term in $\tilde{O}(\cdot)$ depends on $\max_{i\in[m]}\{\frac{1}{\beta_i}\}$.
\end{theorem}

\begin{algorithm}[tb]
\caption{Hedge algorithm}
\label{alg:Hedge}
\begin{algorithmic}
\STATE {\bfseries Input:} a parameter $\eps>0$.
\STATE {\bfseries Initialize:} $\bm{w}_1=\bm{1}\in\mathbb{R}^m$ and $\bm{y}_1=\frac{1}{m}\cdot\bm{w}_1$.
\FOR{$t=1,\dots,T$}
\STATE $\bm{1}.$ Take the action $i_t\sim\bm{y}_t$.
\STATE $\bm{2}.$ Observe the reward vector $\bm{l}_t$ and obtain a reward $l_{i_t, t}$.
\STATE $\bm{3}.$ Update the weight
\[
w_{i,t+1}=w_{i,t}\cdot\exp(-\eps\cdot l_{i,t})
\]
for each $i\in[m]$ and set
\[
y_{i,t+1}=\frac{w_{i,t+1}}{\sum_{i'=1}^{m}w_{i',t+1}}
\]
for each $i\in[m]$.
\ENDFOR
\end{algorithmic}
\end{algorithm}

\section{Missing Proofs for Section \ref{sec:stationary}}
\begin{myproof}[Proof of \Cref{lem:StaDual}]
We aim to prove that
\begin{equation}\label{eqn:121101}
\max_{\bm{\lambda}\geq0}\min_{\tbc_t\in\mathcal{C}} \mathbb{E}_{\tilde{\bm{C}}}[\LInner(\bm{C}, \bm{\lambda})]=\max_{\bm{\lambda}\geq0}\min_{\tbc\in\mathcal{C}} T\cdot\mathbb{E}_{\tbc, \theta\sim P}\left[\bLInner(\tbc, \bm{\lambda}, \theta)\right].
\end{equation}
We denote by $(\bm{\lambda}^*, \tbc^*)$ one optimal solution of the right hand side (RHS) of \eqref{eqn:121101}, where $\tbc$ is allowed to be random. It is clear to see that $(\bm{\lambda}^*, \{\tbc^*_t\}_{\forall t\in[T]})$ with $\tbc^*_t=\tbc^*$ for each $t\in[T]$ is a feasible solution to the left hand side (LHS) of \eqref{eqn:121101}. From the definition of $(\bm{\lambda}^*, \tbc^*)$, it also holds for each $t\in[T]$ that
\[
\tbc^*_t=\tbc^*\in\text{argmin}_{\tbc\in\mathcal{C}}\mathbb{E}_{\tbc, \theta\sim P}\left[\bLInner(\tbc, \bm{\lambda}^*, \theta)\right]
\]
which implies
\[
\bm{C}^*\in\text{argmin}_{\tbc_t\in\mathcal{C}}\mathbb{E}_{\tilde{\bm{C}}}[\LInner(\bm{C}, \bm{\lambda}^*)]
\]
where $\bm{C}^*=(\tbc^*_t)_{t=1}^T$. Therefore, it holds that
\[
\max_{\bm{\lambda}\geq0}\min_{\tbc_t\in\mathcal{C}} \mathbb{E}_{\tilde{\bm{C}}}[\LInner(\tbc, \bm{\lambda}^*)]\geq \max_{\bm{\lambda}\geq0}\min_{\tbc\in\mathcal{C}} T\cdot\mathbb{E}_{\tbc,\theta\sim P}\left[\bLInner(\tbc, \bm{\lambda}^*, \theta)\right].
\]
We now prove the reverse direction. Denote by $(\hat{\bm{\lambda}}, \{\hat{\bm{c}}_t\}_{t=1}^T)$ one optimal solution to the LHS of \eqref{eqn:121101}. Clearly, it holds that for each $t\in[T]$
\[
\hat{\bm{c}}_t\in\text{argmin}_{\tbc\in\mathcal{C}}\mathbb{E}_{\tbc, \theta\sim P}\left[\bLInner(\tbc, \hat{\bm{\lambda}}, \theta)\right].
\]
Then, it is optimal to let $\hat{\bm{c}}_t=\hat{\bm{c}}$ for each $t\in[T]$, for some common $\hat{\bm{c}}\in\mathcal{C}$ such that
\[
\hat{\bm{c}}\in\text{argmin}_{\tbc\in\mathcal{C}}\mathbb{E}_{\tbc, \theta\sim P}\left[\bLInner(\tbc, \hat{\bm{\lambda}}, \theta)\right].
\]
Therefore, $(\hat{\bm{\lambda}}, \hat{\bm{c}})$ is feasible to the RHS of \eqref{eqn:121101} and $\hat{\bm{c}}$ solves the inner minimization problem. We have
\[
\max_{\bm{\lambda}\geq0}\min_{\tbc_t\in\mathcal{C}}\mathbb{E}_{\tilde{\bm{C}}}[\LInner(\tbc, \bm{\lambda}^*)]\leq \max_{\bm{\lambda}\geq0}\min_{\tbc\in\mathcal{C}} T\cdot\mathbb{E}_{\tbc, \theta\sim P}\left[\bLInner(\tbc, \bm{\lambda}^*, \theta)\right]
\]
which completes our proof.
\end{myproof}

\begin{myproof}[Proof of \Cref{lem:ConvexL}]
Fix arbitrary $\bm{\lambda}, \theta$, for any $\bm{c}_1, \bm{c}_2\in\mathcal{C}$ and any $\alpha_1, \alpha_2\geq0$ such that $\alpha_1+\alpha_2=1$, we prove
\[
\alpha_1\cdot\bLInner(\bm{c}_1,\bm{\lambda},\theta)+\alpha_2\cdot\bLInner(\bm{c}_2,\bm{\lambda},\theta)\geq \bLInner(\alpha_1\cdot\bm{c}_1+\alpha_2\cdot\bm{c}_2,\bm{\lambda},\theta).
\]
Now, for $\bLInner(\bm{c}_1,\bm{\lambda},\theta)$, we denote by $\bm{x}_1^*(\theta)$ one optimal solution of the inner minimization problem in the definition of $\bLInner(\bm{c}_1,\bm{\lambda},\theta)$ \eqref{eqn:LISPstationary}. Similarly, for $\bLInner(\bm{c}_2,\bm{\lambda},\theta)$, we denote by $\bm{x}_2^*(\theta)$ one optimal solution of the inner minimization problem in the definition of $\bLInner(\bm{c}_2,\bm{\lambda},\theta)$ \eqref{eqn:LISPstationary}. We then define
\[
\bm{x}^*_3(\theta)=\alpha_1\cdot\bm{x}^*_1(\theta)+\alpha_2\cdot\bm{x}^*_2(\theta),~~\forall \theta.
\]
Under \Cref{assump:main}, from the convexity of $f_{\theta}(\cdot)$ and $\bm{g}_{\theta}(\cdot)$, it holds that
\[\begin{aligned}
&\alpha_1\cdot\left(f_{\theta}(\bm{x}^*_1(\theta))+\sum_{i=1}^{m}\frac{\lambda_i\cdot g_{i,\theta}(\bc_1, \bm{x}^*_1(\theta))}{T\cdot\beta_i}\right)+\alpha_2\cdot\left(f_{\theta}(\bm{x}^*_2(\theta))+\sum_{i=1}^{m}\frac{\lambda_i\cdot g_{i,\theta}(\bc_2, \bm{x}^*_2(\theta))}{T\cdot\beta_i}\right)\\
&\geq f_{\theta}(\bm{x}^*_3(\theta))+\sum_{i=1}^{m}\frac{\lambda_i\cdot g_{i,\theta}(\bc_3, \bm{x}^*_3(\theta))}{T\cdot\beta_i}.
\end{aligned}\]
with $\bc_3=\alpha_1\cdot\bc_1+\alpha_2\cdot\bc_2$.
Given the convexity of $p(\cdot)$, we have
\[\begin{aligned}
&\alpha_1\cdot\bLInner(\bm{c}_1,\bm{\lambda},\theta)+\alpha_2\cdot\bLInner(\bm{c}_2,\bm{\lambda},\theta)\\
=&\alpha_1 p(\bm{c}_1)+\alpha_2 p(\bm{c}_2)-\frac{1}{T}\sum_{i=1}^{m}\lambda_i+\alpha_1\left(f_{\theta}(\bm{x}^*_1(\theta))+\sum_{i=1}^{m}\frac{\lambda_i\cdot g_{i,\theta}(\bc_1, \bm{x}^*_1(\theta))}{T\cdot\beta_i}\right)\\
&+\alpha_2\left(f_{\theta}(\bm{x}^*_2(\theta))+\sum_{i=1}^{m}\frac{\lambda_i\cdot g_{i,\theta}(\bc_2, \bm{x}^*_2(\theta))}{T\cdot\beta_i}\right)\\
\geq& p(\alpha_1\bm{c}_1+\alpha_2\bm{c}_2)+f_{\theta}(\bm{x}^*_3(\theta))+\sum_{i=1}^{m}\frac{\lambda_i\cdot g_{i,\theta}(\bc_3, \bm{x}^*_3(\theta))}{T\cdot\beta_i}\\
\geq& \bLInner(\bc_3,\bm{\lambda},\theta)
\end{aligned}\]
where the last inequality follows from $\bm{x}^*_3(\theta)\in\mathcal{K}(\theta,\bc_3)$, given the exact formulation of $\mathcal{K}(\theta, \bm{c})$ under \Cref{assump:main}.
\end{myproof}

\section{Missing Proofs for Section \ref{sec:adversarial}}

\begin{myproof}[Proof of \Cref{thm:adverLower}]
The proof is modified from the proof of Theorem 2 in \cite{jiang2020online}.
We let $W_T=\mathbb{E}_{\bm{\theta}\sim\bm{P}}[W(\bm{\theta})]$.
We consider a special case of our problem where for any $\bc\in\mathcal{C}$ and any $\theta$, $\mathcal{K}(\theta, \bc)=[0,1]$ (there is no need to decide the first-stage decision). There is only one long-term constraint with target $\beta=\frac{1}{2}$. Moreover, there are three possible values of $\theta$, denoted by $\{\theta^1, \theta^2, \theta^3\}$. We have $f_{\theta^1}(x)=-x$, $f_{\theta^2}(x)=-\left(1+\frac{W_T}{T}\right)x$, $f_{\theta^3}(x)=-\left(1-\frac{W_T}{T}\right)x$ and $g_{\theta^1}(x)=g_{\theta^2}(x)=g_{\theta^3}(x)=x$ (only one long-term constraint). The true distribution is $P_t=\theta^1$ with probability 1 for each $t\in[T]$, and the problem with respect to the true distributions can be described below in \eqref{newappendixeg0}.
\begin{align}
   \min \ \ &  -x_1-...-x_{\frac{T}{2}}-x_{\frac{T}{2}+1}-...-x_{T}  \label{newappendixeg0} \\
    \text{s.t. }\ & x_1+...+x_{\frac{T}{2}}+x_{\frac{T}{2}+1}+...+x_{T} \leq \frac{T}{2}\nonumber \\
    & 0 \leq x_t \leq 1\ \text{ for } t=1,...,T. \nonumber
\end{align}
Now we consider the following two possible adversarial corruptions. The first possible corruption, given in \eqref{newappendixeg3}, is that the distribution $P^c_t=\theta^2$ for $t=\frac{T}{2}+1,\dots,T$. The second possible corruption, given in \eqref{newappendixeg4}, is that the distribution $P^c_t=\theta^3$ for $t=\frac{T}{2}+1,\dots,T$.
\begin{align}
   \min \ \ &  -x_1-...-x_{\frac{T}{2}}-\left(1+\frac{W_T}{T}\right)x_{\frac{T}{2}+1}-...-\left(1+\frac{W_T}{T}\right)x_{T}  \label{newappendixeg3} \\
    \text{s.t. }\ & x_1+...+x_{\frac{T}{2}}+x_{\frac{T}{2}+1}+...+x_{T} \le \frac{T}{2}\nonumber \\
    & 0 \leq x_t \leq 1\ \text{ for } t=1,...,T. \nonumber \\
   \min \ \ &  -x_1-...-x_{\frac{T}{2}} -\left(1-\frac{W_T}{T}\right)x_{\frac{T}{2}+1}-...-\left(1-\frac{W_T}{T}\right)x_{T} \label{newappendixeg4} \\
    \text{s.t. }\ & x_1+...+x_{\frac{T}{2}}+x_{\frac{T}{2}+1}+...+x_{T} \leq \frac{T}{2} \nonumber\\
    & 0 \leq x_t \leq 1\ \text{ for } t=1,...,T.\nonumber
\end{align}
For any online policy $\pi$, denote by $x^1_t(\pi)$ the decision of the policy $\pi$ at period $t$ under corruption scenario given in \eqref{newappendixeg3} and denote by $x^2_t(\pi)$ the decision of the policy $\pi$ at period $t$ under corruption scenario \eqref{newappendixeg4}. Further define $T_1(\pi)$ (resp. $T_2(\pi)$) as the expected capacity consumption of policy $\pi$ under corruption scenario \eqref{newappendixeg3} (resp. corruption scenario \eqref{newappendixeg4}) during the first $\frac{T}{2}$ time periods:
\[
T_1(\pi)=\mathbb{E}\left[\sum_{t=1}^{\frac{T}{2}}x^1_t(\pi)\right] \text{~~~and~~~} T_2(\pi)=\mathbb{E}\left[\sum_{t=1}^{\frac{T}{2}}x^2_t(\pi)\right]
\]
Then, we have that
\[
\ALG_T^1(\pi)=-\frac{T+W_T}{2}+\frac{W_T}{T}\cdot T_1(\pi)\text{~~~and~~~}\ALG_T^2(\pi)=-\frac{T-W_T}{2}-\frac{W_T}{T}\cdot T_2(\pi)
\]
where $\ALG_T^1(\pi)$ (resp. $\ALG_T^2(\pi)$) denotes the expected reward collected by policy $\pi$ on scenario \eqref{newappendixeg3} (resp. scenario \eqref{newappendixeg4}). It is clear to see that the optimal policy $\pi^*$ who is aware of $P^c_t$ for each $t\in[T]$ can achieve an objective value
\[
\ALG_T^1(\pi^*)=-\frac{T+W_T}{2}\text{~~~and~~~}\ALG_T^2(\pi^*)=-\frac{T}{2}.
\]
Thus, the regret of policy $\pi$ on scenario \eqref{newappendixeg3} and \eqref{newappendixeg4} are $\frac{W_T}{T}\cdot T_1(\pi)$ and $W_T-\frac{W_T}{T}\cdot T_2(\pi)$ respectively. Further note that since the implementation of policy $\pi$ at each time period should be independent of future realizations, and more importantly, should independent of corruptions in the future, we must have $T_1(\pi)=T_2(\pi)$ (during the first $\frac{T}{2}$ periods, the information for $\pi$ is the same for both scenarios \eqref{newappendixeg3} and \eqref{newappendixeg4}). Thus, we have that
\[
\text{Reg}_T(\pi)\geq\max\left\{\frac{W_T}{T}\cdot T_1(\pi),W_T-\frac{W_T}{T}\cdot T_1(\pi) \right\}\geq\frac{W_T}{2}=\Omega(W_T)
\]
which completes our proof.
\end{myproof}

\begin{myproof}[Proof of \Cref{thm:CorruptRegretISP}]
The proof follows a similar procedure as the proof of \Cref{thm:RegretISP}.
Denote by $\tau$ the time period that \Cref{alg:DAL} is terminated. There must be a constraint $i'\in[m]$ such that
\begin{equation}\label{eqn:121702}
\sum_{t=1}^{\tau} g_{i', \theta^c_t}(\bc_t, \bm{x}_t)\geq T\cdot \beta_{i'}.
\end{equation}
Otherwise, we can assume without loss of generality that there exists a \textit{dummy} constraint $i'$ such that $g_{i',\theta}(\bc_t, \bm{x})=\beta_{i'}=\alpha$, for arbitrary $\alpha\in(0,1)$, for any $\theta$ and $\bc, \bm{x}$. In this case, we can set $\tau=T$.

We denote by $(\bm{\lambda}^*, \tbc^*)$ one \textit{saddle-point} optimal solution to
\[
\max_{\bm{\lambda}\geq0}\min_{\tbc\in\mathcal{C}} \mathbb{E}_{\tbc^*, \theta\sim P^c}\left[\bLInner(\tbc, \bm{\lambda}, \theta)\right]=\min_{\tbc\in\mathcal{C}}\max_{\bm{\lambda}\geq0} \mathbb{E}_{\tbc^*, \theta\sim P^c}\left[\bLInner(\tbc, \bm{\lambda}, \theta)\right]
\]
where $P^c$ denotes the uniform mixture of the distributions of $\{\theta^c_t\}$ for $t=1$ to $\tau$ and the equality follows from the concavity over $\bm{\lambda}$ and the convexity over $\bm{c}$ proved in \Cref{lem:ConvexL}.
We have
\[
\sum_{t=1}^{\tau}\bLInner(\bm{c}_t, \mu\cdot\bm{e}_{i_t}, \theta^c_t)\leq \sum_{t=1}^{\tau}\mathbb{E}_{\tbc^*}\left[\bLInner(\tbc^*, \mu\cdot\bm{e}_{i_t}, \theta^c_t)\right]+\text{Reg}_1(\tau, \bm{\theta}^c)
\]
following regret bound of $\ALG_1$. Then, it holds that
\begin{align}
\mathbb{E}_{\bm{\theta}\sim\bm{P}}\left[ \sum_{t=1}^{\tau}\bLInner(\bm{c}_t, \mu\cdot\bm{e}_{i_t}, \theta^c_t) \right]&\leq \mathbb{E}_{\tbc^*, \bm{\theta}\sim\bm{P}}\left[ \sum_{t=1}^{\tau}\bLInner(\tbc^*, \mu\cdot\bm{e}_{i_t}, \theta^c_t) \right]+\mathbb{E}_{\bm{\theta}\sim\bm{P}}\left[\text{Reg}_1(\tau, \bm{\theta}^c)\right] \label{eqn:121701}\\
&\leq \tau\cdot \mathbb{E}_{\tbc^*, \theta\sim P^c}\left[ \bLInner(\tbc^*, \bm{\lambda}^*, \theta) \right]+\mathbb{E}_{\bm{\theta}\sim\bm{P}}\left[\text{Reg}_1(\tau, \bm{\theta}^c)\right] \nonumber\\
&\leq \tau\cdot \mathbb{E}_{\tbc^*, \theta\sim P}\left[ \bLInner(\tbc^*, \bm{\lambda}^*, \theta) \right]+\mathbb{E}_{\bm{\theta}\sim\bm{P}}\left[\text{Reg}_1(\tau, \bm{\theta}^c)\right]+O(\mathbb{E}_{\bm{\theta}\sim\bm{P}}[W(\bm{\theta})]). \nonumber
\end{align}
On the other hand, for any $i\in[m]$, we have
\[
\sum_{t=1}^{\tau}\bLInner(\bm{c}_t, \mu\cdot\bm{e}_{i_t}, \theta^c_t)\geq \sum_{t=1}^{\tau}\bLInner_i(\bm{c}_t, \mu\cdot\bm{e}_{i_t}, \theta^c_t)-\text{Reg}_2(\tau, \bm{\theta}^c)
\]
following the regret bound of $\ALG_2$ (holds for arbitrary $\bm{\lambda}=\mu\cdot\bm{e}_i$). We now set $i=i'$ and we have
\[\begin{aligned}
\sum_{t=1}^{\tau}\bLInner(\bm{c}_t, \mu\cdot\bm{e}_{i_t}, \theta^c_t)&\geq \sum_{t=1}^{\tau}\bLInner_{i'}(\bm{c}_t, \mu\cdot\bm{e}_{i_t}, \theta^c_t)-\text{Reg}_2(\tau, \bm{\theta}^c)\\
&=\sum_{t=1}^{\tau} \left(p(\bm{c}_t)+f_{\theta^c_t}(\bm{x}_t)\right)-\mu\cdot \frac{\tau}{T}+\sum_{t=1}^{\tau}\frac{\mu\cdot g_{i',\theta^c_t}(\bc_t, \bm{x}_t)}{T\cdot\beta_{i'}}-\text{Reg}_2(\tau, \bm{\theta}^c)\\
&\geq \sum_{t=1}^{\tau} \left(p(\bm{c}_t)+f_{\theta^c_t}(\bm{x}_t)\right)+\mu\cdot\frac{T-\tau}{T}-\text{Reg}_2(\tau, \bm{\theta}^c)
\end{aligned}\]
where the last inequality follows from \eqref{eqn:121702}. Then, it holds that
\begin{equation}\label{eqn:121703}
\mathbb{E}_{\bm{\theta}\sim\bm{P}}\left[ \sum_{t=1}^{\tau}\bLInner(\bm{c}_t, \mu\cdot\bm{e}_{i_t}, \theta^c_t) \right]\geq \mathbb{E}_{\bm{\theta}\sim\bm{P}}\left[ \sum_{t=1}^{\tau} \left(p(\bm{c}_t)+f_{\theta^c_t}(\bm{x}_t)\right) \right]+\mu\cdot\frac{T-\tau}{T}-\mathbb{E}_{\bm{\theta}\sim\bm{P}}\left[ \text{Reg}_2(\tau, \bm{\theta}^c) \right].
\end{equation}
Combining \eqref{eqn:121701} and \eqref{eqn:121703}, we have
\[\begin{aligned}
\mathbb{E}_{\bm{\theta}\sim\bm{P}}\left[ \sum_{t=1}^{\tau} \left(p(\bm{c}_t)+f_{\theta^c_t}(\bm{x}_t)\right) \right]\leq& -\mu\cdot\frac{T-\tau}{T}+\tau\cdot \mathbb{E}_{\tbc^*, \theta\sim P}\left[ \bLInner(\tbc^*, \bm{\lambda}^*, \theta) \right]+O(\mathbb{E}_{\bm{\theta}\sim\bm{P}}[W(\bm{\theta})])\\
&+\mathbb{E}_{\bm{\theta}\sim\bm{P}}\left[\text{Reg}_1(\tau, \bm{\theta})\right]+\mathbb{E}_{\bm{\theta}\sim\bm{P}}\left[ \text{Reg}_2(\tau, \bm{\theta}) \right].
\end{aligned}\]
From the boundedness conditions in \Cref{assump:main}, we have
\[
\mathbb{E}_{\tbc^*, \theta\sim P}\left[\bLInner(\tbc^*, \bm{\lambda}^*, \theta)\right]=\frac{1}{T}\cdot\OPTInner\geq -1
\]
which implies that
\[
-\mu\cdot\frac{T-\tau}{T}\leq (T-\tau)\cdot \mathbb{E}_{\tbc^*, \theta\sim P}\left[\bLInner(\tbc^*, \bm{\lambda}^*, \theta)\right]
\]
when $\mu=T$. Therefore, we have
\begin{align}
\mathbb{E}_{\bm{\theta}\sim\bm{P}}\left[ \sum_{t=1}^{\tau} \left(p(\bm{c}_t)+f_{\theta^c_t}(\bm{x}_t)\right) \right]\leq& T\cdot \mathbb{E}_{\tbc^*, \theta\sim P}\left[ \bLInner(\tbc^*, \bm{\lambda}^*, \theta) \right]+O(\mathbb{E}_{\bm{\theta}\sim\bm{P}}[W(\bm{\theta})])+\mathbb{E}_{\bm{\theta}\sim\bm{P}}\left[\text{Reg}_1(T, \bm{\theta})\right] \nonumber\\
&+\mathbb{E}_{\bm{\theta}\sim\bm{P}}\left[ \text{Reg}_2(T, \bm{\theta}) \right]\label{eqn:121704}\\
\leq& \OPT^c+O(\mathbb{E}_{\bm{\theta}\sim\bm{P}}[W(\bm{\theta})])+\mathbb{E}_{\bm{\theta}\sim\bm{P}}\left[\text{Reg}_1(T, \bm{\theta})\right] \nonumber\\
&+\mathbb{E}_{\bm{\theta}\sim\bm{P}}\left[ \text{Reg}_2(T, \bm{\theta}) \right]. \nonumber
\end{align}
where $\OPT^c$ denotes the value of the optimal policy with adversarial corruptions. It is clear to see that $|\OPT^c-\OPT|\leq O(\mathbb{E}_{\bm{\theta}\sim\bm{P}}[W(\bm{\theta})])$.
Using \Cref{thm:OGDregret} and \Cref{thm:Hedgeregret} to bound $\mathbb{E}_{\bm{\theta}\sim\bm{P}}\left[\text{Reg}_1(T, \bm{\theta})\right]$ and $\mathbb{E}_{\bm{\theta}\sim\bm{P}}\left[ \text{Reg}_2(T, \bm{\theta}) \right]$, we have
\[
\mathbb{E}_{\bm{\theta}\sim\bm{P}}\left[ \sum_{t=1}^{\tau} \left(p(\bm{c}_t)+f_{\theta^c_t}(\bm{x}_t)\right) \right]\leq \OPT^c+O((G+F)\cdot\sqrt{T})+O(\sqrt{T\cdot\log m})+O(\mathbb{E}_{\bm{\theta}\sim\bm{P}}[W(\bm{\theta})])
\]
which completes our proof of \eqref{eqn:121801}.
\end{myproof}

\section{Missing Proofs for Section \ref{sec:nonstationary}}

\begin{myproof}[Proof of \Cref{thm:priorLower}]
The proof is modified from the proof of \Cref{thm:adverLower}.
We consider a special case of our problem where for any $\bc\in\mathcal{C}$ and any $\theta$, $\mathcal{K}(\theta, \bc)=[0,1]$ (there is no need to decide the first-stage decision). There is only one long-term constraint with target $\beta=\frac{1}{2}$. Moreover, there are three possible values of $\theta$, denoted by $\{\theta^1, \theta^2, \theta^3\}$. We have $f_{\theta^1}(x)=-x$, $f_{\theta^2}(x)=-\left(1+\frac{W_T}{T}\right)x$, $f_{\theta^3}(x)=-\left(1-\frac{W_T}{T}\right)x$ and $g_{\theta^1}(x)=g_{\theta^2}(x)=g_{\theta^3}(x)=x$ (only one long-term constraint). The prior estimate is $\hat{P}_t=\theta^1$ with probability 1 for each $t\in[T]$, and the problem with respect to the prior estimates can be described below in \eqref{appendixeg0}.
\begin{align}
   \min \ \ &  -x_1-...-x_{\frac{T}{2}}-x_{\frac{T}{2}+1}-...-x_{T}  \label{appendixeg0} \\
    \text{s.t. }\ & x_1+...+x_{\frac{T}{2}}+x_{\frac{T}{2}+1}+...+x_{T} \leq \frac{T}{2}\nonumber \\
    & 0 \leq x_t \leq 1\ \text{ for } t=1,...,T. \nonumber
\end{align}
Now we consider the following two possible true distributions. The first possible true scenario, given in \eqref{appendixeg3}, is that the distribution  $P_t=\theta^1$ for $t=1,\dots,\frac{T}{2}$ and $P_t=\theta^2$ for $t=\frac{T}{2}+1,\dots,T$. The second possible true scenario, given in \eqref{appendixeg4}, is that the distribution $P_t=\theta^1$ for $t=1,\dots,\frac{T}{2}$ and $P^c_t=\theta^3$ for $t=\frac{T}{2}+1,\dots,T$.
\begin{align}
   \min \ \ &  -x_1-...-x_{\frac{T}{2}}-\left(1+\frac{W_T}{T}\right)x_{\frac{T}{2}+1}-...-\left(1+\frac{W_T}{T}\right)x_{T}  \label{appendixeg3} \\
    \text{s.t. }\ & x_1+...+x_{\frac{T}{2}}+x_{\frac{T}{2}+1}+...+x_{T} \le \frac{T}{2}\nonumber \\
    & 0 \leq x_t \leq 1\ \text{ for } t=1,...,T. \nonumber \\
   \min \ \ &  -x_1-...-x_{\frac{T}{2}} -\left(1-\frac{W_T}{T}\right)x_{\frac{T}{2}+1}-...-\left(1-\frac{W_T}{T}\right)x_{T} \label{appendixeg4} \\
    \text{s.t. }\ & x_1+...+x_{\frac{T}{2}}+x_{\frac{T}{2}+1}+...+x_{T} \leq \frac{T}{2} \nonumber\\
    & 0 \leq x_t \leq 1\ \text{ for } t=1,...,T.\nonumber
\end{align}
For any online policy $\pi$, denote by $x^1_t(\pi)$ the decision of the policy $\pi$ at period $t$ under the true scenario given in \eqref{appendixeg3} and denote by $x^2_t(\pi)$ the decision of the policy $\pi$ at period $t$ under the true scenario \eqref{appendixeg4}. Further define $T_1(\pi)$ (resp. $T_2(\pi)$) as the expected capacity consumption of policy $\pi$ under the true scenario \eqref{appendixeg3} (resp. true scenario \eqref{appendixeg4}) during the first $\frac{T}{2}$ time periods:
\[
T_1(\pi)=\mathbb{E}\left[\sum_{t=1}^{\frac{T}{2}}x^1_t(\pi)\right] \text{~~~and~~~} T_2(\pi)=\mathbb{E}\left[\sum_{t=1}^{\frac{T}{2}}x^2_t(\pi)\right]
\]
Then, we have that
\[
\ALG_T^1(\pi)=-\frac{T+W_T}{2}+\frac{W_T}{T}\cdot T_1(\pi)\text{~~~and~~~}\ALG_T^2(\pi)=-\frac{T-W_T}{2}-\frac{W_T}{T}\cdot T_2(\pi)
\]
where $\ALG_T^1(\pi)$ (resp. $\ALG_T^2(\pi)$) denotes the expected reward collected by policy $\pi$ on scenario \eqref{appendixeg3} (resp. scenario \eqref{appendixeg4}). It is clear to see that the optimal policy $\pi^*$ who is aware of $P_t$ for each $t\in[T]$ can achieve an objective value
\[
\ALG_T^1(\pi^*)=-\frac{T+W_T}{2}\text{~~~and~~~}\ALG_T^2(\pi^*)=-\frac{T}{2}.
\]
Thus, the regret of policy $\pi$ on scenario \eqref{appendixeg3} and \eqref{appendixeg4} are $\frac{W_T}{T}\cdot T_1(\pi)$ and $W_T-\frac{W_T}{T}\cdot T_2(\pi)$ respectively. Further note that since the implementation of policy $\pi$ at each time period should be independent of future realizations, we must have $T_1(\pi)=T_2(\pi)$ (during the first $\frac{T}{2}$ periods, the information for $\pi$ is the same for both scenarios \eqref{appendixeg3} and \eqref{appendixeg4}). Thus, we have that
\[
\text{Reg}_T(\pi)\geq\max\left\{\frac{W_T}{T}\cdot T_1(\pi),W_T-\frac{W_T}{T}\cdot T_1(\pi) \right\}\geq\frac{W_T}{2}=\Omega(W_T)
\]
which completes our proof.
\end{myproof}

\begin{myproof}[Proof of \Cref{lem:decompose}]
Denote by $(\hat{\bm{\lambda}}^*, (\hbc^*_t)_{t=1}^T)$ the optimal solution to
\[
\max_{\bm{\lambda}\geq0}\min_{\bm{c}_t\in\mathcal{C}} \hLInner(\bm{C}, \bm{\lambda}),
\]
used in the definition \eqref{eqn:121902}.
we now show that $(\hat{\bm{\lambda}}^*, \hbc^*_t)$ is an optimal solution to
\[
\max_{\bm{\lambda}\geq0}\min_{\bm{c}_t\in\mathcal{C}} \hLInner_t(\bm{c}_t, \bm{\lambda})
\]
for each $t\in[T]$, which would help to complete our proof of \eqref{eqn:121904}. We first define
\[
L_t(\bm{\lambda})=\min_{\bm{c}_t\in\mathcal{C}} \hLInner_t(\bm{c}_t, \bm{\lambda}),
\]
as a function over $\bm{\lambda}$ for each $t\in[T]$. Then, it holds that
\begin{equation}\label{eqn:121905}
\nabla L_t(\hat{\bm{\lambda}}^*)=\nabla \hLInner_t(\hbc^*_t, \hat{\bm{\lambda}}^*)=\left( -\frac{\hat{\beta}_{i,t}}{T\beta_i}+\mathbb{E}_{\theta\sim\hat{P}_t}\left[ \frac{g_{i,\theta}(\hbc^*_t, \hbx^*_t(\theta))}{T\beta_i} \right] \right)_{\forall i\in[m]}=\bm{0}
\end{equation}
The first equality of \eqref{eqn:121905} follows from the fact that
\begin{equation}\label{eqn:121906}
\hbc^*_t\in\text{argmin}_{\bc\in\mathcal{C}}\mathbb{E}_{\theta\sim\hat{P}_t}\left[p(\bc)+
\min_{\bx\in\mathcal{K}(\theta,\bc)}f_{\theta}(\bx)+\sum_{i=1}^{m}\frac{\hat{\lambda}^*_i\cdot g_{i,\theta}(\bc, \bx)}{T\cdot\beta_i}\right]
\end{equation}
since $(\hat{\bm{\lambda}}^*, (\hbc^*_t)_{t=1}^T)$ is an optimal solution to $\max_{\bm{\lambda}\geq0}\min_{\bm{c}_t\in\mathcal{C}} \hLInner(\bm{C}, \bm{\lambda})$. The last equality of \eqref{eqn:121905} follows from the definition of $\hat{\beta}_{i,t}$ in \eqref{eqn:121902}. Therefore, combining \eqref{eqn:121905} and \eqref{eqn:121906}, we know that $(\hat{\bm{\lambda}}^*, \hbc^*_t)$ is an optimal solution to
$\max_{\bm{\lambda}\geq0}\min_{\bm{c}_t\in\mathcal{C}} \hLInner_t(\bm{c}_t, \bm{\lambda})$
for each $t\in[T]$.

We now prove \eqref{eqn:121904}. It is sufficient to prove that
\begin{equation}\label{eqn:121907}
  \hLInner(\hat{\bm{C}}^*, \hat{\bm{\lambda}}^*)=\sum_{t=1}^{T} \hLInner_t(\hbc^*_t, \hat{\bm{\lambda}}^*)
\end{equation}
where $\hat{\bm{C}}^*=(\hbc^*_t)_{t=1}^T$. We define an index set $\mathcal{I}=\{i\in[m]: \hat{\lambda}^*_i>0\}$. Clearly, for each $i\in\mathcal{I}$, the optimality of $\hat{\bm{\lambda}}^*$ would require that
\[
\nabla_{\lambda_i}\bLInner(\hat{\bm{C}}^*, \hat{\bm{\lambda}}^*)=-1+\frac{1}{T\beta_i}\sum_{t=1}^{T}\mathbb{E}_{\theta\sim\hat{P}_t}\left[ g_{i,\theta}(\hbc_t^*, \hbx^*_t(\theta)) \right]=0
\]
which implies $\sum_{t=1}^{T}\hat{\beta}_{i,t}=T\cdot\beta$. Therefore, we would have
\[
\sum_{i=1}^{m}\sum_{t=1}^{T}\frac{\hat{\lambda}^*_{i,t}\hat{\beta}_i}{T\beta_i}=\sum_{i=1}^{m}\hat{\lambda}^*_i.
\]
which completes our proof of \eqref{eqn:121908} and therefore \eqref{eqn:121904}.
\end{myproof}

\section{Missing Proofs for Appendix \ref{sec:extensions}}\label{sec:pfextensions}

\begin{myproof}[Proof of \Cref{thm:DiscreteC}]
The proof follows a similar procedure as the proof of \Cref{thm:CorruptRegretISP}.
Denote by $\tau$ the time period that \Cref{alg:DAL} is terminated. There must be a constraint $i'\in[m]$ such that
\begin{equation}\label{eqn:012502}
\sum_{t=1}^{\tau} g_{i', \theta^c_t}(\bc_t, \bm{x}_t)\geq T\cdot \beta_{i'}.
\end{equation}
Otherwise, we can assume without loss of generality that there exists a \textit{dummy} constraint $i'$ such that $g_{i',\theta}(\bc_t, \bm{x})=\beta_{i'}=\alpha$, for arbitrary $\alpha\in(0,1)$, for any $\theta$ and $\bc, \bm{x}$. In this case, we can set $\tau=T$.

We denote by $(\bm{\lambda}^*, \tbc^*)$ one \textit{saddle-point} optimal solution to
\begin{equation}\label{eqn:012506}
\max_{\bm{\lambda}\geq0}\min_{\tbc\in\mathcal{C}} \mathbb{E}_{\tbc^*, \theta\sim P^c}\left[\bLInner(\tbc, \bm{\lambda}, \theta)\right]=\min_{\tbc\in\mathcal{C}}\max_{\bm{\lambda}\geq0} \mathbb{E}_{\tbc^*, \theta\sim P^c}\left[\bLInner(\tbc, \bm{\lambda}, \theta)\right]
\end{equation}
where $P^c$ denotes the uniform mixture of the distributions of $\{\theta^c_t\}$ for $t=1$ to $\tau$. We note that since $\tbc$ is a distribution over $K$ discrete points $\{\bm{c}^1,\dots,\bm{c}^K\}$, $\tbc$ can be denoted by $\bm{p}_{\tbc}\in\mathbb{R}^m$ with $0\leq\bm{p}_{\tbc}$ and $\|\bm{p}_{\tbc}\|_1=1$. Then, we have
\[
\mathbb{E}_{\tbc^*, \theta\sim P^c}\left[\bLInner(\tbc, \bm{\lambda}, \theta)\right]=\langle \bm{p}_{\tbc}, \bar{\bm{L}} \rangle
\]
with $\bar{\bm{L}}=( \bLInner(\bc^k, \bm{\lambda}, \theta) )_{k=1}^K\in\mathbb{R}^K$. Therefore, $\mathbb{E}_{\tbc^*, \theta\sim P^c}\left[\bLInner(\tbc, \bm{\lambda}, \theta)\right]$ can be regarded as a convex function over $\tbc$ and is clearly a concave function over $\bm{\lambda}$. The equality \eqref{eqn:012506} follows from the Sion's minimax theorem \citep{sion1958general}.

We have
\[
\sum_{t=1}^{\tau}\bLInner(\bm{c}_t, \mu\cdot\bm{e}_{i_t}, \theta^c_t)\leq \sum_{t=1}^{\tau}\mathbb{E}_{\tbc^*}\left[\bLInner(\tbc^*, \mu\cdot\bm{e}_{i_t}, \theta^c_t)\right]+\text{Reg}_1(\tau, \bm{\theta}^c)
\]
following regret bound of $\ALG_1$. Then, it holds that
\begin{align}
\mathbb{E}_{\bm{\theta}\sim\bm{P}}\left[ \sum_{t=1}^{\tau}\bLInner(\bm{c}_t, \mu\cdot\bm{e}_{i_t}, \theta^c_t) \right]&\leq \mathbb{E}_{\tbc^*, \bm{\theta}\sim\bm{P}}\left[ \sum_{t=1}^{\tau}\bLInner(\tbc^*, \mu\cdot\bm{e}_{i_t}, \theta^c_t) \right]+\mathbb{E}_{\bm{\theta}\sim\bm{P}}\left[\text{Reg}_1(\tau, \bm{\theta}^c)\right] \label{eqn:012503}\\
&\leq \tau\cdot \mathbb{E}_{\tbc^*, \theta\sim P^c}\left[ \bLInner(\tbc^*, \bm{\lambda}^*, \theta) \right]+\mathbb{E}_{\bm{\theta}\sim\bm{P}}\left[\text{Reg}_1(\tau, \bm{\theta}^c)\right] \nonumber\\
&\leq \tau\cdot \mathbb{E}_{\tbc^*, \theta\sim P}\left[ \bLInner(\tbc^*, \bm{\lambda}^*, \theta) \right]+\mathbb{E}_{\bm{\theta}\sim\bm{P}}\left[\text{Reg}_1(\tau, \bm{\theta}^c)\right]+O(\mathbb{E}_{\bm{\theta}\sim\bm{P}}[W(\bm{\theta})]). \nonumber
\end{align}
On the other hand, for any $i\in[m]$, we have
\[
\sum_{t=1}^{\tau}\bLInner(\bm{c}_t, \mu\cdot\bm{e}_{i_t}, \theta^c_t)\geq \sum_{t=1}^{\tau}\bLInner_i(\bm{c}_t, \mu\cdot\bm{e}_{i_t}, \theta^c_t)-\text{Reg}_2(\tau, \bm{\theta}^c)
\]
following the regret bound of $\ALG_2$ (holds for arbitrary $\bm{\lambda}=\mu\cdot\bm{e}_i$). We now set $i=i'$ and we have
\[\begin{aligned}
\sum_{t=1}^{\tau}\bLInner(\bm{c}_t, \mu\cdot\bm{e}_{i_t}, \theta^c_t)&\geq \sum_{t=1}^{\tau}\bLInner_{i'}(\bm{c}_t, \mu\cdot\bm{e}_{i_t}, \theta^c_t)-\text{Reg}_2(\tau, \bm{\theta}^c)\\
&=\sum_{t=1}^{\tau} \left(p(\bm{c}_t)+f_{\theta^c_t}(\bm{x}_t)\right)-\mu\cdot \frac{\tau}{T}+\sum_{t=1}^{\tau}\frac{\mu\cdot g_{i',\theta^c_t}(\bc_t, \bm{x}_t)}{T\cdot\beta_{i'}}-\text{Reg}_2(\tau, \bm{\theta}^c)\\
&\geq \sum_{t=1}^{\tau} \left(p(\bm{c}_t)+f_{\theta^c_t}(\bm{x}_t)\right)+\mu\cdot\frac{T-\tau}{T}-\text{Reg}_2(\tau, \bm{\theta}^c)
\end{aligned}\]
where the last inequality follows from \eqref{eqn:012502}. Then, it holds that
\begin{equation}\label{eqn:012504}
\mathbb{E}_{\bm{\theta}\sim\bm{P}}\left[ \sum_{t=1}^{\tau}\bLInner(\bm{c}_t, \mu\cdot\bm{e}_{i_t}, \theta^c_t) \right]\geq \mathbb{E}_{\bm{\theta}\sim\bm{P}}\left[ \sum_{t=1}^{\tau} \left(p(\bm{c}_t)+f_{\theta^c_t}(\bm{x}_t)\right) \right]+\mu\cdot\frac{T-\tau}{T}-\mathbb{E}_{\bm{\theta}\sim\bm{P}}\left[ \text{Reg}_2(\tau, \bm{\theta}^c) \right].
\end{equation}
Combining \eqref{eqn:012503} and \eqref{eqn:012504}, we have
\[\begin{aligned}
\mathbb{E}_{\bm{\theta}\sim\bm{P}}\left[ \sum_{t=1}^{\tau} \left(p(\bm{c}_t)+f_{\theta^c_t}(\bm{x}_t)\right) \right]\leq& -\mu\cdot\frac{T-\tau}{T}+\tau\cdot \mathbb{E}_{\tbc^*, \theta\sim P}\left[ \bLInner(\tbc^*, \bm{\lambda}^*, \theta) \right]+O(\mathbb{E}_{\bm{\theta}\sim\bm{P}}[W(\bm{\theta})])\\
&+\mathbb{E}_{\bm{\theta}\sim\bm{P}}\left[\text{Reg}_1(\tau, \bm{\theta})\right]+\mathbb{E}_{\bm{\theta}\sim\bm{P}}\left[ \text{Reg}_2(\tau, \bm{\theta}) \right].
\end{aligned}\]
From the boundedness condition c in \Cref{assump:main}, we have
\[
\mathbb{E}_{\tbc^*, \theta\sim P}\left[\bLInner(\tbc^*, \bm{\lambda}^*, \theta)\right]=\frac{1}{T}\cdot\OPTInner\geq -1
\]
which implies that
\[
-\mu\cdot\frac{T-\tau}{T}\leq (T-\tau)\cdot \mathbb{E}_{\tbc^*, \theta\sim P}\left[\bLInner(\tbc^*, \bm{\lambda}^*, \theta)\right]
\]
when $\mu=T$. Therefore, we have
\begin{align}
\mathbb{E}_{\bm{\theta}\sim\bm{P}}\left[ \sum_{t=1}^{\tau} \left(p(\bm{c}_t)+f_{\theta^c_t}(\bm{x}_t)\right) \right]\leq& T\cdot \mathbb{E}_{\tbc^*, \theta\sim P}\left[ \bLInner(\tbc^*, \bm{\lambda}^*, \theta) \right]+O(\mathbb{E}_{\bm{\theta}\sim\bm{P}}[W(\bm{\theta})])+\mathbb{E}_{\bm{\theta}\sim\bm{P}}\left[\text{Reg}_1(T, \bm{\theta})\right] \nonumber\\
&+\mathbb{E}_{\bm{\theta}\sim\bm{P}}\left[ \text{Reg}_2(T, \bm{\theta}) \right]\label{eqn:012505}\\
\leq& \OPT^c+O(\mathbb{E}_{\bm{\theta}\sim\bm{P}}[W(\bm{\theta})])+\mathbb{E}_{\bm{\theta}\sim\bm{P}}\left[\text{Reg}_1(T, \bm{\theta})\right] \nonumber\\
&+\mathbb{E}_{\bm{\theta}\sim\bm{P}}\left[ \text{Reg}_2(T, \bm{\theta}) \right]. \nonumber
\end{align}
where $\OPT^c$ denotes the value of the optimal policy with adversarial corruptions. It is clear to see that $|\OPT^c-\OPT|\leq O(\mathbb{E}_{\bm{\theta}\sim\bm{P}}[W(\bm{\theta})])$.
Using Theorem 3.1 in \cite{auer2002nonstochastic} to bound $\mathbb{E}_{\bm{\theta}\sim\bm{P}}\left[\text{Reg}_1(T, \bm{\theta})\right]$ and \Cref{thm:Hedgeregret} to bound $\mathbb{E}_{\bm{\theta}\sim\bm{P}}\left[ \text{Reg}_2(T, \bm{\theta}) \right]$, we have
\[
\mathbb{E}_{\bm{\theta}\sim\bm{P}}\left[ \sum_{t=1}^{\tau} \left(p(\bm{c}_t)+f_{\theta^c_t}(\bm{x}_t)\right) \right]\leq \OPT^c+O((G+F)\cdot\sqrt{T})+O(\sqrt{T\cdot\log m})+O(\mathbb{E}_{\bm{\theta}\sim\bm{P}}[W(\bm{\theta})])
\]
which completes our proof of \eqref{eqn:012501}.
\end{myproof}

\begin{myproof}[Proof of \Cref{thm:RegretOSP}]
From the formulation \eqref{eqn:LOSPstationary}, the dual variable $\bm{\lambda}$ is scaled by $T$. Therefore, we know that $\mu=\alpha\cdot T$ for some constant $\alpha>0$.

We assume without loss of generality that there always exists $i'\in[m]$ such that
\begin{equation}\label{eqn:121602}
\sum_{t=1}^{T} g_{i', \theta_t}(\bc_t, \bm{x}_t)\leq T\cdot \beta_{i'}.
\end{equation}
In fact, let there be a \textit{dummy} constraint $i'$ such that $g_{i',\theta}(\bc, \bm{x})=\beta_{i'}=\alpha$, for arbitrary $\alpha\in(0,1)$, for any $\theta$ and $\bc, \bm{x}$. Then, \eqref{eqn:121602} holds.

We denote by $(\bm{\lambda}^*, \tbc^*)$ one \textit{saddle-point} optimal solution to
\[
\max_{\bm{\lambda}\geq0}\min_{\tbc\in\mathcal{C}} \mathbb{E}_{\tbc, \theta\sim P}\left[\bLOuter(\tbc, \bm{\lambda}, \theta)\right]=\min_{\tbc\in\mathcal{C}}\max_{\bm{\lambda}\geq0} \mathbb{E}_{\tbc, \theta\sim P}\left[\bLOuter(\tbc, \bm{\lambda}, \theta)\right].
\]
We have
\[
\sum_{t=1}^{T}\bLOuter(\bm{c}_t, \mu\cdot\bm{e}_{i_t}, \theta_t)\leq \sum_{t=1}^{T}\mathbb{E}_{\tbc^*}\left[\bLOuter(\tbc^*, \mu\cdot\bm{e}_{i_t}, \theta_t)\right]+\text{Reg}_1(T, \bm{\theta})
\]
following regret bound of $\ALG_1$. Then, it holds that
\begin{equation}\label{eqn:121601}
\begin{aligned}
\mathbb{E}_{\bm{\theta}\sim\bm{P}}\left[ \sum_{t=1}^{T}\bLOuter(\bm{c}_t, \mu\cdot\bm{e}_{i_t}, \theta_t) \right]&\leq \mathbb{E}_{\tbc^*, \bm{\theta}\sim\bm{P}}\left[ \sum_{t=1}^{T}\bLOuter(\tbc^*, \mu\cdot\bm{e}_{i_t}, \theta_t) \right]+\mathbb{E}_{\bm{\theta}\sim\bm{P}}\left[\text{Reg}_1(T, \bm{\theta})\right]\\
&\leq \tau\cdot \mathbb{E}_{\tbc^*, \theta\sim P}\left[ \bLOuter(\tbc^*, \bm{\lambda}^*, \theta) \right]+\mathbb{E}_{\bm{\theta}\sim\bm{P}}\left[\text{Reg}_1(T, \bm{\theta})\right].
\end{aligned}
\end{equation}
On the other hand, for any $i\in[m]$, we have
\[
\sum_{t=1}^{T}\bLOuter(\bm{c}_t, \mu\cdot\bm{e}_{i_t}, \theta_t)\geq \sum_{t=1}^{T}\bLOuter_i(\bm{c}_t, \mu\cdot\bm{e}_{i_t}, \theta_t)-\text{Reg}_2(T, \bm{\theta})
\]
following the regret bound of $\ALG_2$ in (holds for arbitrary $\bm{\lambda}=\mu\cdot\bm{e}_i$). We now denote by
\[
i^*=\text{argmax}_{i\in[m]}\{ \beta_i-\frac{1}{T}\cdot\sum_{t=1}^{T}g_{i,\theta_t}(\bc_t, \bm{x}_t) \}.
\]
We also denote by
\[
d_T(\mathcal{A}, \bm{\theta})=\max_{i\in[m]}\{ \beta_i-\frac{1}{T}\cdot\sum_{t=1}^{T}g_{i,\theta_t}(\bc_t, \bm{x}_t) \}
\]
as the distance away from the target set $\mathcal{A}$. Following \eqref{eqn:121602}, we always have $d_T(\mathcal{A}, \bm{\theta})\geq0$.
We now set $i=i^*$ and we have
\[\begin{aligned}
\sum_{t=1}^{T}\bLOuter(\bm{c}_t, \mu\cdot\bm{e}_{i_t}, \theta_t)&\geq \sum_{t=1}^{T}\bLInner_{i^*}(\bm{c}_t, \mu\cdot\bm{e}_{i_t}, \theta_t)-\text{Reg}_2(T, \bm{\theta})\\
&=\sum_{t=1}^{T} \left(p(\bm{c}_t)+f_{\theta_t}(\bm{x}_t)\right)+\mu-\sum_{t=1}^{T}\frac{\mu\cdot g_{i^*,\theta_t}(\bc_t, \bm{x}_t)}{T\cdot\beta_{i^*}}-\text{Reg}_2(T, \bm{\theta})\\
&\geq \sum_{t=1}^{T} \left(p(\bm{c}_t)+f_{\theta_t}(\bm{x}_t)\right)+\frac{\mu}{\beta_{i^*}}\cdot d_T(\mathcal{A}, \bm{\theta})-\text{Reg}_2(T, \bm{\theta})
\end{aligned}\]
where the last inequality follows from \eqref{eqn:121602}. Then, it holds that
\begin{equation}\label{eqn:121603}
\begin{aligned}
\mathbb{E}_{\bm{\theta}\sim\bm{P}}\left[ \sum_{t=1}^{T}\bLOuter(\bm{c}_t, \mu\cdot\bm{e}_{i_t}, \theta_t) \right]\geq & \mathbb{E}_{\bm{\theta}\sim\bm{P}}\left[ \sum_{t=1}^{T} \left(p(\bm{c}_t)+f_{\theta_t}(\bm{x}_t)\right) \right]\\
&+\frac{\mu}{\beta_{\max}}\cdot \mathbb{E}_{\bm{\theta}\sim\bm{P}} [d_T(\mathcal{A}, \bm{\theta})]-\mathbb{E}_{\bm{\theta}\sim\bm{P}}\left[ \text{Reg}_2(T, \bm{\theta}) \right].
\end{aligned}
\end{equation}
where $\beta_{\max}=\max_{i\in[m]}\{\beta_i\}$.
Combining \eqref{eqn:121601} and \eqref{eqn:121603}, we have
\begin{equation}\label{eqn:121604}
\begin{aligned}
\mathbb{E}_{\bm{\theta}\sim\bm{P}}\left[ \sum_{t=1}^{T} \left(p(\bm{c}_t)+f_{\theta_t}(\bm{x}_t)\right) \right]+\frac{\mu}{\beta_{\max}}\cdot \mathbb{E}_{\bm{\theta}\sim\bm{P}} [d_T(\mathcal{A}, \bm{\theta})]\leq& T\cdot \mathbb{E}_{\theta\sim P}\left[ \bLOuter(\bm{c}^*, \bm{\lambda}^*, \theta) \right]\\
&+\mathbb{E}_{\bm{\theta}\sim\bm{P}}\left[\text{Reg}_1(T, \bm{\theta})\right]+\mathbb{E}_{\bm{\theta}\sim\bm{P}}\left[ \text{Reg}_2(T, \bm{\theta}) \right].
\end{aligned}
\end{equation}
From the non-negativity of $d_T(\mathcal{A}, \bm{\theta})$, we have
\[
\mathbb{E}_{\bm{\theta}\sim\bm{P}}\left[ \sum_{t=1}^{T} \left(p(\bm{c}_t)+f_{\theta_t}(\bm{x}_t)\right) \right]\leq T\cdot \mathbb{E}_{\theta\sim P}\left[ \bLOuter(\bm{c}^*, \bm{\lambda}^*, \theta) \right]+\mathbb{E}_{\bm{\theta}\sim\bm{P}}\left[\text{Reg}_1(T, \bm{\theta})\right]+\mathbb{E}_{\bm{\theta}\sim\bm{P}}\left[ \text{Reg}_2(T, \bm{\theta}) \right].
\]
Using \Cref{thm:OGDregret} and \Cref{thm:Hedgeregret} to bound $\mathbb{E}_{\bm{\theta}\sim\bm{P}}\left[\text{Reg}_1(T, \bm{\theta})\right]$ and $\mathbb{E}_{\bm{\theta}\sim\bm{P}}\left[ \text{Reg}_2(T, \bm{\theta}) \right]$, we have
\[
\mathbb{E}_{\bm{\theta}\sim\bm{P}}\left[ \sum_{t=1}^{T} \left(p(\bm{c}_t)+f_{\theta_t}(\bm{x}_t)\right) \right]\leq T\cdot \mathbb{E}_{\theta\sim P}\left[ \bLOuter(\bm{c}^*, \bm{\lambda}^*, \theta) \right]+O((G+F)\cdot\sqrt{T})+O(\sqrt{T\cdot\log m})
\]
which completes our proof of \eqref{eqn:121501}.

We now prove \eqref{eqn:121502}. We note that $(\bm{c}_t, \bm{x}_t)_{t=1}^T$ defines a feasible solution to
\begin{align}
 \OPT^{\delta}= \min & \sum_{t=1}^{T}\mathbb{E}_{\bm{c}_t, \bm{x}_t, \theta_t}\left[p(\bm{c}_t)+f_{\theta_t}(\bm{x}_t)\right] \label{lp:OuterA}\\
  \mbox{s.t.} & \frac{1}{T}\cdot\sum_{t=1}^{T}\mathbb{E}_{\bc_t, \bm{x}_t, \theta_t}[\bm{g}_{\theta_t}(\bc_t, \bm{x}_t)]\geq\bm{\beta}-\delta \nonumber\\
  & \bm{x}_t\in\mathcal{K}(\theta_t, \bm{c}_t), \bm{c}_t\in\mathcal{C}, \forall t.\nonumber
\end{align}
with $\delta=\mathbb{E}_{\bm{\theta}\sim\bm{P}}[d_T(\mathcal{A}, \bm{\theta})]$. If we regard $\OPT^{\delta}$ as a function over $\delta$, then $\OPT^{\delta}$ is clearly a convex function over $\delta$, where the proof follows the same spirit as the proof of \Cref{lem:ConvexL}. Moreover, note that
\[
\frac{d\OPT^{\delta=0}}{d\delta}=\|\bm{\lambda^*}\|_1\leq\mu.
\]
We have
\[
\OPTOuter=\OPT^{0}\leq \OPT^{\delta}+  \mu\cdot \delta
\]
for any $\delta\geq0$. Therefore, it holds that
\begin{equation}\label{eqn:121605}
\begin{aligned}
T\cdot \mathbb{E}_{\theta\sim P}\left[ \bLOuter(\bm{c}^*, \bm{\lambda}^*, \theta) \right]&=\OPTOuter \leq \OPT^{\mathbb{E}_{\bm{\theta}\sim\bm{P}}[d_T(\mathcal{A},\bm{\theta})]}+ \mu\cdot \mathbb{E}_{\bm{\theta}\sim\bm{P}}[d_T(\mathcal{A},\bm{\theta})]\\
&\leq \sum_{t=1}^{T} \mathbb{E}_{\bm{\theta}\sim\bm{P}}\left[p(\bm{c}_t)+f_{\theta_t}(\bm{x}_t)+\mu\cdot d_T(\mathcal{A},\bm{\theta})\right]
\end{aligned}
\end{equation}
Plugging \eqref{eqn:121605} into \eqref{eqn:121604}, we have
\[
\mu\cdot\left(\frac{1}{\beta_{\max}}-1\right)\cdot \mathbb{E}_{\bm{\theta}\sim\bm{P}}[ d_T(\mathcal{A},\bm{\theta})]\leq \mathbb{E}_{\bm{\theta}\sim\bm{P}}\left[\text{Reg}_1(T, \bm{\theta})\right]+\mathbb{E}_{\bm{\theta}\sim\bm{P}}\left[ \text{Reg}_2(T, \bm{\theta}) \right].
\]
Our proof of \eqref{eqn:121502} is completed by using \Cref{thm:OGDregret} and \Cref{thm:Hedgeregret} to bound $\mathbb{E}_{\bm{\theta}\sim\bm{P}}\left[\text{Reg}_1(T, \bm{\theta})\right]$ and $\mathbb{E}_{\bm{\theta}\sim\bm{P}}\left[ \text{Reg}_2(T, \bm{\theta}) \right]$.
\end{myproof}

\end{APPENDICES}

\end{document}